\documentclass[manuscript]{acmart}
\AtBeginDocument{%
  }

\setcopyright{acmlicensed}
\copyrightyear{2026}
\acmYear{2026}
\acmDOI{XXXXXXX.XXXXXXX}
\acmConference[Conference acronym 'XX]{Make sure to enter the correct
  conference title from your rights confirmation email}{June 03--05,
  2018}{Woodstock, NY}
\acmISBN{978-1-4503-XXXX-X/2018/06}

\usepackage{listings}
\usepackage{xcolor}
\usepackage{tcolorbox}
\tcbuselibrary{breakable, skins}
\usepackage{subcaption}
\usepackage{graphicx}

\usepackage{booktabs}
\usepackage{multirow}
\usepackage{colortbl}
\usepackage{array}

\newcommand{\modelicon}[1]{%
  \raisebox{-2pt}{\includegraphics[height=10pt]{logos/#1}}%
}

\definecolor{myblue}{HTML}{4C72B0}

\definecolor{typeA}{HTML}{FADBDD}   
\definecolor{typeB}{HTML}{FDEEE2}   
\definecolor{typeC}{HTML}{D8EDEA}   
\definecolor{typeD}{HTML}{DDE7ED}   
\definecolor{typeE}{HTML}{EDE1FA}   
\definecolor{textA}{HTML}{E63946}
\definecolor{textB}{HTML}{F4A261}
\definecolor{textC}{HTML}{2A9D8F}
\definecolor{textD}{HTML}{457B9D}
\definecolor{textE}{HTML}{9B5DE5}

\usepackage{soul}
\renewcommand{\hl}[1]{#1}

\lstdefinestyle{jsonStyle}{
    language=json,
    basicstyle=\ttfamily\small,
    numbers=left,
    numberstyle=\tiny\color{gray},
    stepnumber=1,
    numbersep=8pt,
    backgroundcolor=\color{gray!10},
    showspaces=false,
    showstringspaces=false,
    showtabs=false,
    frame=single,
    rulecolor=\color{gray!40},
    tabsize=2,
    captionpos=b,
    breaklines=true,
    breakatwhitespace=false,
    stringstyle=\color{teal},
    keywordstyle=\color{blue},
    commentstyle=\color{gray},
}

\tcbset{
  promptbox/.style={
    enhanced,
    breakable,
    colback=white,
    colframe=myblue,
    colbacktitle=myblue,
    coltitle=white,
    fonttitle=\bfseries\small,
    title=#1,
    left=6pt, right=6pt, top=4pt, bottom=4pt,
    fontupper=\ttfamily\small,
  }
}

\definecolor{jsonkey}{RGB}{0,0,180}
\definecolor{jsonstring}{RGB}{163,21,21}
\definecolor{jsonnumber}{RGB}{128,0,128}
\definecolor{jsonbackground}{RGB}{245,245,245}

\lstdefinelanguage{json}{
    basicstyle=\ttfamily\small,
    numbers=left,
    numberstyle=\tiny\color{gray},
    stepnumber=1,
    numbersep=10pt,
    showstringspaces=false,
    breaklines=true,
    frame=single,
    backgroundcolor=\color{jsonbackground},
    string=[s]{"}{"},
    stringstyle=\color{jsonstring},
    comment=[l]{//},
    commentstyle=\color{gray},
    moredelim=[s][\color{jsonkey}]{\{}{\:},
    moredelim=[s][\color{jsonkey}]{,}{\:},
    literate=
     *{0}{{{\color{jsonnumber}0}}}{1}
      {1}{{{\color{jsonnumber}1}}}{1}
      {2}{{{\color{jsonnumber}2}}}{1}
      {3}{{{\color{jsonnumber}3}}}{1}
      {4}{{{\color{jsonnumber}4}}}{1}
      {5}{{{\color{jsonnumber}5}}}{1}
      {6}{{{\color{jsonnumber}6}}}{1}
      {7}{{{\color{jsonnumber}7}}}{1}
      {8}{{{\color{jsonnumber}8}}}{1}
      {9}{{{\color{jsonnumber}9}}}{1}
}
\begin{document}

\title[\texttt{GTBench}: A Benchmark for Evaluating LLMs in graph Theory ]{\texttt{GTBench}: A Curriculum-Grounded Benchmark for Evaluating LLMs as Mathematical Research Assistants in Graph Theory}

\author{Noujoud Nader, Ibrahem Aljabea}
\email{nnader@lsu.edu}
\orcid{1234-5678-9012}
\email{Ibrahem.Aljabea@lsu.edu}
\affiliation{%
  \institution{Louisiana State University}
  \city{Baton Rouge}
  \state{LA}
  \country{USA}
}
  
\author{Patrick Diehl}
\affiliation{%
  \institution{Los Alamos National Laboratory}
  \city{Los Alamos National Laboratory}
    \state{NM}
  \country{USA}}
\email{diehlpk@lanl.gov}
\orcid{1234-5678-9012}

\author{Deepti Gupta}
\affiliation{%
  \institution{Texas A\&M-Central Texas}
  \city{College Station}
  \state{TX}
  \country{USA}
}

\renewcommand{\shortauthors}{Nader et al.}

\begin{abstract}
Large language models (LLMs) are increasingly used as
self-study assistant in technical disciplines, yet their
reliability as mathematical reasoning assistants remains
poorly understood. We introduce \texttt{GTBench}, a
curriculum-grounded benchmark for evaluating LLMs as
mathematical research assistants in graph theory,
comprising 63 problems organized into three groups of
increasing difficulty: undergraduate definitions and
basic properties (Group~1), algorithm tracing and
structural reasoning (Group~2), and graduate-level proof
construction (Group~3). Problems are sourced from
verified academic materials including Diestel's
\emph{Graph Theory}. We evaluate five frontier
models --- \texttt{GPT-5}, \texttt{Claude Sonnet~4.6}, \texttt{Gemini~2.5
Flash-Lite}, \texttt{Llama~3.3~70B}, and \texttt{Mistral Large~3} --- under
zero-shot and chain-of-thought prompting, using exact-match
and LLM-as-judge evaluation for Groups~1 and~2, and a
hybrid human expert and LLM-as-judge protocol for Group~3.
Our results reveal a pronounced performance hierarchy:
\texttt{GPT-5} approaches ceiling on Group~1 (95.8\% zero-shot)
and maintains meaningful accuracy on graduate proofs
(82\%), while all other models degrade substantially with
difficulty, with \texttt{Llama} achieving 0\% under human
evaluation on Group~3 zero-shot. 
Failure mode analysis shows that \textit{correct
algorithm, wrong execution} errors dominate Groups~1
and~2, while Group~3 additionally surfaces
\textit{incomplete reasoning} failures and reveals
systematic disagreement between human evaluators and
the automated judge, particularly on verbose or
near-complete proofs ($\kappa = 0.48$--$0.83$ across
human pairs). \texttt{GTBench} provides the first
curriculum-grounded evaluation framework for
graph-theoretic reasoning in LLMs, with direct
implications for the governance of AI tools in
mathematical education and scientific research.
\end{abstract}




\maketitle

\section{Introduction}
\label{Intro}
Large language models (LLMs) have demonstrated remarkable performance across a wide range of reasoning tasks, from commonsense question answering to mathematical problem solving and code generation~\cite{cobbe2021gsm8k,hendrycks2021math,diehl2024evaluating,nader2025llm,diehl2025llm,mhatre2026can}. These advances have driven growing interest in understanding the extent and limits of LLM reasoning, particularly in domains requiring structured, multi-step thinking. Among these, mathematics has emerged as a central proving ground, offering problems with well-defined correct answers and clear gradations of difficulty.
However, the mathematical reasoning literature has largely concentrated on numerical, algebraic, and formal proof settings. Graph-theoretic reasoning occupies a unique position in this field, it is at once elementary in its definitions and deeply challenging in its proofs, requiring the kind of structural intuition that develops only through sustained practice. And it is precisely because of this richness that it serves as an ideal stress-test for LLMs: can these models move beyond pattern-matched answers to genuinely reason about relational structure, combinatorial properties, and formal proof construction? This question is not merely theoretical. Researchers and students increasingly rely on LLMs as reasoning assistants in their daily work~\cite{nader2025llm,diehl2025llmhpc}, and the gap between assumed and actual capability in a domain as foundational as graph theory carries real consequences for how these tools are trusted and adopted.

Graph theory constitutes a core branch of discrete mathematics and theoretical computer science, offering significant contributions to algorithm design, network analysis, combinatorial optimization, complexity theory, and data structures \cite{nader2015classification, nader2015pregnancy}. Problems in graph theory demand a distinctive form of reasoning: one that is simultaneously visual and abstract, requiring the manipulation of relational structures, the application of algorithmic procedures, and the justification of structural properties. Unlike arithmetic or algebraic reasoning, graph-theoretic problem solving often requires a model to track the state of a combinatorial object across multiple reasoning steps, recognize structural patterns, and apply theorems whose conditions must be carefully verified. These characteristics make graph theory an especially informative and challenging domain for probing the reasoning capabilities of LLMs.

Graph-theoretic reasoning has not been systematically evaluated in the LLM benchmarking literature. Existing benchmarks either focus on numerical and algebraic reasoning, as in MATH~\cite{hendrycks2021math}, GSM8K~\cite{cobbe2021gsm8k}, and GHOSTS~\cite{frieder2023ghosts} or on formal, machine-verifiable proof generation, as in LeanDojo~\cite{yang2023leandojo} and MiniF2F~\cite{zheng2021minif2f}. Research-level benchmarks such as FrontierMath~\cite{glazer2024frontiermath}, HLE~\cite{center2026benchmark}, LemmaBench~\cite{peyronnet2026lemmabenchliveresearchlevelbenchmark}, and RealMath~\cite{zhang2025realmathcontinuousbenchmarkevaluating} target expert-level mathematical difficulty but do not specifically isolate graph-theoretic reasoning as an evaluation domain. While recent work has begun to probe whether LLMs can solve graph problems expressed in natural language~\cite{wang2023can, fatemi2024talk}, a systematic, curriculum-grounded benchmark spanning the full range of graph theory instruction has remained absent from the literature. \hl{The growing use of LLMs for mathematical reasoning raises a governance challenge: users need evidence of when model outputs can be reliably used, particularly because mathematical errors may be difficult to detect when reasoning appears coherent or plausible. This requires evaluating reliability across different levels of reasoning to inform appropriate use and human oversight. Existing mathematical benchmarks provide limited evidence for these needs because they generally do not isolate graph-theoretic reasoning or distinguish foundational knowledge from algorithmic execution and open-ended mathematical justification.} 




\hl{To address this limitation, we introduce \texttt{\textbf{GTBench}}, a curriculum-grounded benchmark that organizes graph-theoretic problems into progressively more demanding levels of reasoning, enabling model capabilities and limitations to be characterized beyond aggregate performance. By providing evidence about reliability across these reasoning levels, GTBench can inform decisions about appropriate use, human verification, and oversight when LLMs are employed as mathematical reasoning assistants.} The current study focuses on Group~1 (undergraduate introductory) and Group~2 (undergraduate intermediate), with Group~3  covering graduate-level content.\\

\textit{Group~1} problems assess foundational knowledge: familiarity with standard graph families such as complete graphs $K_n$, cycles $C_n$, paths $P_n$, wheel graphs $W_n$, hypercubes $Q_d$, and complete bipartite graphs $K_{r,s}$, as well as degree sequences, subgraph operations, complement graphs, isomorphism, and elementary counting arguments. These problems are predominantly definitional and combinatorial in nature, and their answers are compact enough to be evaluated by exact match.\\

\textit{Group~2} problems step up in complexity, requiring the application of graph algorithms and structural reasoning about properties that emerge from connectivity and traversal. Topics include breadth-first search (BFS) and depth-first search (DFS), cut vertices and bridges, Eulerian and Hamiltonian conditions, tree characterizations, and spanning tree algorithms such as Kruskal's and Prim's. These problems correspond to the algorithmic core of a standard undergraduate discrete mathematics or graph theory course, requiring models to execute multi-step procedures, track intermediate states, and produce verifiable reasoning at each step.\\

\textit{Group~3} problems move beyond standard algorithmic reasoning and require deeper mathematical justification, proof construction, and comparative evaluation of multiple solution strategies. These problems often involve advanced graph-theoretic concepts such as graph coloring, planarity, matching, network flows, spectral properties, NP-complete graph problems, and complexity-based reasoning. In contrast to Groups~1 and~2, Group~3 emphasizes open-ended reasoning where multiple valid approaches may exist, and partial correctness must be carefully assessed. Because of this complexity, evaluation in Group~3 combines {human expert judges} and {LLM-as-judge} evaluation to ensure reliable scoring and consistency.

\hl{The deployment of LLMs as learning and research companions in
scientific disciplines raises questions that extend beyond raw
accuracy: how consistent are their outputs across repeated
queries, what types of errors do they make, and at what level
of task complexity does their reliability degrade to the point
where unverified use becomes methodologically unsound? GTBench
is designed to answer these questions systematically for graph
theory, providing the curriculum-grounded empirical evidence
needed to characterise LLM behaviour across the full range of
difficulty encountered in a standard graduate programme.}
\noindent Our main contributions are as follows:
\begin{itemize}
  \item We introduce \textbf{\texttt{\textbf{GTBench}}}, the first curriculum-grounded benchmark for evaluating graph-theoretic reasoning in LLMs, framed explicitly as an evaluation of LLMs as mathematical research assistants and organized into three groups aligned with the standard progression of graph theory instruction.
 
  \item We conduct a systematic empirical evaluation of leading LLMs, providing detailed analysis of model strengths and failure modes across a range of graph-theoretic problem types.
 
  \item We design and validate a \textbf{hybrid evaluation methodology} combining LLM-as-judge scoring and human expert evaluation.
 
  \item We release \texttt{\textbf{GTBench}} publicly to support reproducible research and to serve as a foundation for future evaluation of LLM reasoning in discrete mathematics and theoretical computer science.
\end{itemize}

\paragraph{Paper Structure} The remainder of this paper is organized as follows. Section~\ref{sec:related:work} reviews the related work. The benchmark design is presented in Section~\ref{sec:benchmark:design}. The filtering procedure is described in Section~\ref{sec:filtering:procedure}. The experimental methodology is outlined in Section~\ref{sec:evaluation:methodology}, followed by the failure mode taxonomy in Section~\ref{sec:failure_taxonomy}. The results are discussed in Section~\ref{sec:results:discussion}. Section~\ref{sec:conclusion} concludes the paper, outlining directions for future work.

\section{Related Work}
\label{sec:related:work}
Efforts to benchmark LLM mathematical reasoning span a wide difficulty spectrum. At the school and competition level, GSM8K~\cite{cobbe2021gsm8k} covers grade-school arithmetic, MATH~\cite{hendrycks2021math} targets competition-level algebra and geometry, and GHOSTS~\cite{frieder2023ghosts} probes undergraduate problem solving. 
At the research frontier: \textit{(i)} LemmaBench~\cite{peyronnet2026lemmabenchliveresearchlevelbenchmark} used \texttt{Gemini} 2.5 Pro and ChatGPT 4/5, with results judged by both LLMs and humans; \textit{(ii)} RealMath~\cite{zhang2025realmathcontinuousbenchmarkevaluating} draws from arXiv preprints and mathematical forums, using ChatGPT o3/o4, \texttt{Gemini} 2.5 Pro, Claude 3.7, Grok-3, and DeepSeek R1, with results judged by LLMs; \textit{(iii)} FrontierMath~\cite{glazer2024frontiermath} used GPT-4, Claude 3.7, and \texttt{Gemini} 1.5 Pro, with results verified through automated verification; and \textit{(iv)} HLE~\cite{center2026benchmark} uses expert-curated problems to push models to their limits, with the initial publication evaluating GPT-4o/o1 and Claude 3.5. Their leaderboard additionally reports results for GPT o3-mini/o3/o5, Sonnet 3.7/4/4.5, \texttt{Gemini} 2.5 Pro/3.x, Grok 4, DeepSeek R1, and Kimi K2. Table~\ref{tab:llm:summary} summarizes these works. A separate line of work, represented by LeanDojo~\cite{yang2023leandojo} and MiniF2F~\cite{zheng2021minif2f}, focuses on machine-verifiable proof generation within interactive theorem provers. Across all these efforts, problems are drawn primarily from numerical, algebraic, and analytic domains, leaving discrete and graph-theoretic mathematics largely unrepresented.
In this work, the authors~\cite{xie2026core} introduced CORE, a human-verified benchmark to evaluate the semantic reasoning capabilities of LLMs on fundamental static code analysis tasks. The authors~\cite{kulkarni2026evaluating} evaluated how effectively a LLM can support graph-theoretic reasoning by testing it on both a solved problem and an open research problem using a structured mathematical evaluation process. Wang et al.~\cite{wang2025exploring} conducted a comprehensive study of LLMs on graph learning tasks, comparing their reasoning abilities, robustness, and transfer performance with traditional graph learning models across different evaluation settings. In this work, the authors~\cite{liu2025combibench} introduced CombiBench, a formal benchmark and evaluation framework for testing the combinatorial reasoning abilities of LLMs using Lean 4–based mathematical problems. 

Most closely related to our work, \cite{wang2023can} investigates whether LLMs can solve graph problems expressed in natural language, surfacing important limitations in structural reasoning. \hl{They use a random graph generator to generate various graphs. The LLMs were tested on: connectivity, cycles, topological sort, shortest path, maximum flow, bipartite graph matching, Hamilton path, and graph neural networks. These exercises using a graph visualization or graph description were found in the textbooks used in this work}. The work in \cite{fatemi2024talk} \hl{studies how the choice of text encoding for a graph affects LLM performance on basic graph reasoning tasks.}
 However, these works lack a curriculum-grounded structure and do not address the evaluation challenges posed by multi-step procedural problems and proofs.\\

\hl{Note that the size of the items in these benchmarks vary quite a lot. The highest items with 12,500 are in the MATH benchmark}~\cite{hendrycks2021math}. \hl{The following topics are covered: algebra, number theory, calculus, geometry, counting, and probability. GSM8k}~\cite{cobbe2021gsm8k} \hl{includes 8,792 grade-school mathematical problems. The problems range from basic arithmetic; fractions; decimals; percentages; counting and elementary; to combinatorics.} HLE~\cite{center2026benchmark} \hl{includes 2,500 items including algebra and number theory; combinatorics and discrete mathematics; geometry and topology; real, complex, and functional analysis; probability and statistics; mathematical logic; and applied mathematics; and computational mathematics.} RealMath~\cite{zhang2025realmathcontinuousbenchmarkevaluating} \hl{includes 1,286 items from combinatorics; number theory; algebraic and symplectic geometry; probability and information theory; group, representation, and ring theory; 
functional, real, and complex analysis; differential equations; mathematical physics; optimization and control; 
discrete mathematics and algorithms; machine learning and computational complexity; calculus; integration; limits; sequences; and series. These benchmarks are above one thousand items.}

GHOSTS~\cite{frieder2023ghosts} \hl{includes 709 items ranging from functional analysis to topology and probability theory. }LemmaBench~\cite{peyronnet2026lemmabenchliveresearchlevelbenchmark} \hl{has 358 items among the following topics: computer science/information theory, mathematical physics, analysis of partial differential equations, differential geometry, dynamical systems, functional analysis, and many more.} FrontierMath~\cite{glazer2024frontiermath}\hl{includes 338 items from the following topics: number theory; real and complex analysis; algebra and algebraic geometry; combinatorics; category theory; representation theory; topology; differential geometry; partial differential equations; probability theory; mathematical logic; and theoretical computer science.}

\hl{At the smaller end of the spectrum, CombiBench contains 100 problems focused exclusively on combinatorics. In contrast, most other benchmarks cover multiple topics. FrontierMath, for example, contains 338 problems across 12 topics—an average of approximately 28 problems per topic, assuming an even distribution. CombiBench and GTBench therefore stand out for their focus on a single topic. Although they contain fewer problems overall than broader benchmarks, their number of problems per topic is comparable to that of some larger multi-topic benchmarks.}

    

\begin{table*}[tb]
\centering
\caption{Comparison of related benchmarks and \textit{\texttt{\textbf{GTBench}}}
         (this work) across difficulty level, models evaluated,
         data source, and evaluation method. \textbf{Auto.}~=
         automated verification; \textbf{LLM}~= LLM-as-judge;
         \textbf{Human}~= human expert evaluation. \textbf{N} is the number of items in each benchmark.}
\label{tab:llm:summary}
\setlength{\extrarowheight}{3pt}
\resizebox{\textwidth}{!}{%
\begin{tabular}{
  >{\bfseries}p{2.6cm}
  p{2.8cm}
  p{5.cm}
  p{1.8cm}
  p{2.0cm}
  p{0.8cm}
}
\toprule
\textbf{Name}
  & \textbf{Level}
  & \textbf{LLMs evaluated}
  & \textbf{Data source}
  & \textbf{Judges} 
  & \textbf{N}\\
  
\midrule
 
\rowcolor{white}
LemmaBench~\cite{peyronnet2026lemmabenchliveresearchlevelbenchmark}
  & Lemma
  & \texttt{Gemini} 2.5 Pro, GPT-4/5
  & arXiv
  & LLM/Human 
  & 358 \\

GSM8K~\cite{cobbe2021gsm8k}
  & Grade-school mathematical reasoning
  & GPT-3, GPT-3 + Verifier
  & arXiv
  & Human/Verifier 
  & 8,792\\

MATH~\cite{hendrycks2021math}
  & High-school competition mathematics
  & GPT-3, PaLM, GPT-4, Claude, Gemini, Llama-series
  & Authors
  & Exact-match answer checking 
  & 12,500 \\

GHOSTS~\cite{frieder2023ghosts}
  & Graduate-level mathematics
  & ChatGPT, GPT-4
  & Authors
  & Human 
  & 709\\
 
RealMath~\cite{zhang2025realmathcontinuousbenchmarkevaluating}
  & Research-level
  & GPT-o3/o4, \texttt{Gemini} 2.5 Pro, Claude 3.7, Grok-3, DeepSeek R1
  & arXiv
  & LLM 
  & 1,286\\
 

FrontierMath~\cite{glazer2024frontiermath}
  & Unpublished research-level
  & GPT-4, Claude 3.5, \texttt{Gemini} 1.5 Pro
  & Authors
  & Auto 
  & 338\\
 

CombiBench~\cite{liu2025combibench}
  & Combinatorics / Formal reasoning
  & Kimina-Prover, GPT-based models
  & arXiv
  & Fine-Eval and Lean-based evaluation
  & 100 \\

HLE~\cite{center2026benchmark}
  & Expert-level
  & GPT-4o/o1, Claude 3.5--4.5, \texttt{Gemini} 2.5 Pro/3.x, Grok~4,
    DeepSeek R1, Kimi K2
  & Community
  & Auto. 
  & 2,500\\
 
\midrule
 
\rowcolor{typeD}
\texttt{\textbf{\texttt{\textbf{GTBench}}}}
\newline
\small\textit{(This work)}
  & Graduate exercises
  & \texttt{GPT-5}, \texttt{Gemini} 2.5 Flash-Lite,
    Llama 3.3 70B, Mistral Large, \texttt{Claude Sonnet} 4.6
  & Textbook
  & LLM/Human 
  & 63\\
 
\bottomrule
\end{tabular}%
}
\end{table*}

\section{Benchmark Design}
\label{sec:benchmark:design}
In this study, we introduce \texttt{\textbf{\texttt{\textbf{GTBench}}}}, a curriculum-grounded benchmark for
evaluating the graph-theoretic reasoning capabilities of LLMs. This section describes the design
principles underlying \texttt{\textbf{GTBench}}, the sources from which problems were
drawn, and the three groups that constitute the current study. We organize the problems into three groups corresponding to the
standard progression of graph theory instruction from first-year
undergraduate to graduate level. Table~\ref{tab:tier_overview}
summarizes the groups structure. Group~1 (introductory) problems test familiarity
with the foundational vocabulary and properties of graphs including standard
graph families ($K_n$, $C_n$, $P_n$, $W_n$, $Q_d$, $K_{r,s}$),
degree sequences, subgraph operations, complement graphs, isomorphism,
and elementary counting arguments.
Group~2 (intermediate) problems require the
application of graph algorithms and the reasoning about structural
properties that emerge from connectivity and traversal. Topics include
BFS and DFS, cut vertices and bridges, Eulerian and Hamiltonian
conditions, tree characterisations, and spanning tree algorithms.
These correspond to the second half of a standard graph theory course
and require the researcher to execute multi-step procedures and justify
their outputs.  Group 3 (graduate — proof construction) problems require the construction of formal mathematical arguments, the application of graduate-level theorems, and in several cases the identification of counterexamples to disprove stated claims. Topics span matching theory (Hall's theorem, Tutte's theorem), connectivity (Menger's theorem, 2-connected and 3-connected graphs), planarity (outerplanar graphs, self-dual plane graphs), graph coloring, flows, and cycle structure. These problems correspond to the content of a graduate graph theory course and demand that the model not only identify the correct theorem or technique but produce a logically sound and complete proof — a qualitatively harder task than algorithm tracing or definitional recall. Unlike Groups~1 and~2, where answers are evaluated
by exact match and LLM-as-judge against a ground
truth, Group~3 problems are evaluated by both human
expert judges and an LLM-as-judge using a structured
rubric (Section~\ref{sec:evaluation:methodology}),
given the inherent complexity of graduate-level proof
construction and the need for reliable human
validation at this level.

\subsection{Source}
\texttt{\textbf{GTBench}} problems  were drawn from two primary academic
sources, chosen for their scholarly credibility, unambiguous
provenance, and well-documented difficulty structure.

\begin{table}[ht]
\caption{\texttt{\textbf{\texttt{\textbf{GTBench}}}} groups structure.}
\centering
\label{tab:tier_overview}
\begin{tabular}{clp{5.2cm}cc}
\toprule
\textbf{Group \#} & \textbf{Level} & \textbf{Description}
              & \textbf{Problems} \\
\midrule
1 & Definitions \& Basics
  & Definitions, basic graph families, degree arguments, counting, isomorphism
  & 31   \\[4pt]
2 & Algorithms \& Structures
  & Algorithm tracing (BFS/DFS), walks, connectivity, Eulerian \& Hamiltonian graphs
  & 21  \\[4pt]
3 & Proof Construction
  & Theorem proving, matching, planarity, colouring, flows
  & 11\\

\midrule
\multicolumn{3}{r}{\textbf{Total}} & \textbf{63} & \\
\bottomrule
\end{tabular}
\end{table}

 \subsubsection{Source A: Diestel's \emph{Graph Theory} and CMU~21-484
Solutions.}
The first source is the exercise set of Diestel's \emph{Graph Theory}~\cite{diestel},
the standard graduate reference textbook in the field, now in its fifth edition.
Ground truth answers for Group~1 Diestel exercises were taken directly from
the homework solution sets of Carnegie Mellon University's course 21-484
Graph Theory~\cite{cmu_hw}, which provide faculty-endorsed reference proofs
for a subset of Diestel's Chapter~1 exercises. Ground truth answers for
Group~3 problems were provided by a graph theory expert at Louisiana State
University (LSU). 

\subsubsection{Source B: UPC Mathematics~1 Graph Theory Problem Set.}
The second source is the official problem set for the Mathematics~1
course (Graph Theory module) taught at the Facultat
d'Inform\`{a}tica de Barcelona, Universitat Polit\`{e}cnica de
Catalunya (UPC), for the academic year 2025--2026~\cite{upc_problems, upc_solutions}.
This collection was assembled by faculty members of the Departament de
Matem\`{a}tiques and covers the standard undergraduate graph theory
curriculum: basic graph families and properties, subgraphs and
operations, walks and connectivity, Eulerian and Hamiltonian graphs,
and trees. It is particularly well-suited to Groups~1 and~2 of
\texttt{\textbf{\texttt{\textbf{GTBench}}}} because it is structured as a teaching resource with clearly
differentiated exercise types, ranging from definitional recall and
adjacency matrix construction at the introductory level to non-trivial
proof exercises requiring induction or counting arguments. 
 
All solutions of the three groups were
reviewed and validated by graph theory expert prior to inclusion in \texttt{\textbf{\texttt{\textbf{GTBench}}}},
ensuring that the reference answers meet the standards of mathematical
rigour expected at each curricular level.
 
\section{Filtering procedure}
\label{sec:filtering:procedure}
Starting from the full exercise sets of both sources, we applied the
following exclusion criteria in order:
\begin{enumerate}
  \item \textit{Figure dependency}: exclude any problem whose statement
        requires reading or interpreting a graph drawing.
  \item \textit{Ambiguity}: exclude problems with multiple defensible
        interpretations or that depend on conventions not specified in
        the problem statement.
  \item \textit{Ground truth unavailability}: exclude problems for
        which no verified reference answer could be established.
  \item \textit{Group boundary}: exclude problems whose difficulty
        clearly falls outside the intended group range (\emph{e.g.}, problems
        requiring graduate-level machinery in a Group~1 context).
\end{enumerate}
After filtering, 13 problems were retained from Source~A and
18 from Source~B, yielding 31 problems for Group~1. Group~2 draws
exclusively from Source~B (Chapters~2--4), retaining 21 problems
whose ground truth answers were verified against the official solution
file. Group~3 draws from graduate-level proof construction
problems rooted in Diestel's \emph{Graph Theory}~\cite{diestel},
retaining 11 problems evaluated by human expert judges
and an LLM-as-judge
given the proof-construction nature of the tasks (Table~\ref{tab:tier_overview}). 

Each problem in \texttt{\textbf{\texttt{\textbf{GTBench}}}} is stored as a JSON object with an evaluation
type of either \textsf{exact\_match} or \textsf{proof}. Figure~\ref{fig:json_example} shows a
representative Group~1 problem as stored in \texttt{\textbf{\texttt{\textbf{GTBench}}}}. Problems of
type \textsf{exact\_match} are those where the model's answer can be
compared directly to the ground truth string or numerical value, used
for problems with a unique, fully specified correct answer (\emph{e.g.}, the
number of spanning trees of $K_{2,r}$, or the Eulerian condition for
$Q_n$). Problems in  Group~2 problems and 13 of the 31 Group~1 problems are
of this type. Problems of type \textsf{proof} require the model to
produce a natural-language or semi-formal mathematical argument, which
cannot be assessed by string comparison alone. For Group~1, the 18
proof-type problems are evaluated using an LLM-as-judge approach with
a structured rubric (Section~\ref{sec:eval_method}). For Group~3, all 11 problems are evaluated by both
human graph theory experts and an LLM-as-judge,
since graduate-level proof construction requires
human validation to ensure reliability and
mathematical rigour.

\begin{figure}[ht]
\begin{lstlisting}[style=jsonStyle, caption={A representative \texttt{\textbf{\texttt{\textbf{GTBench}}}}
Group~1 problem entry (JSON format).}, label={fig:json_example}]
{
  "id": "T1_009",
  "tier": 1,
  "source": "UPC Graph Theory, Exercise 1.16",
  "topic": "Parity argument",
  "question": "Prove that if a graph is regular of odd
               degree, then it has even order.",
  "ground_truth": "Let G be r-regular with r odd and order
                   n. By the handshaking lemma, the sum of
                   all degrees equals 2m (even). The sum
                   equals n*r. Since r is odd, n*r is even
                   if and only if n is even.",
  "requires_figure": false,
  "evaluation_type": "proof"
}
\end{lstlisting}
\end{figure}

\section{Experimental Methodology}
\label{sec:evaluation:methodology}
\subsection{Models Evaluated}
We evaluate five large language models spanning the major families of
frontier and open-weight systems currently available to researchers.
The models were selected to provide broad coverage across developers
(Anthropic, Google, OpenAI, Meta, Mistral), architectural approaches
(proprietary dense, open-weight dense, mixture-of-experts), and
capability tiers (frontier flagship, efficient mid-tier, open-weight).
Table~\ref{tab:models} summarises the five models and their key
characteristics. We use \texttt{GPT-4o} as the LLM judge rather than \texttt{GPT-5} to
avoid self-serving bias, as using the same model to
both generate and evaluate answers could systematically
favour its own outputs~\cite{zheng2023judging}.

\begin{table}[ht]
\centering
\caption{LLMs evaluated in \texttt{\textbf{\texttt{\textbf{GTBench}}}}. All models were accessed via
         their respective commercial APIs using the exact model
         strings listed.}
\label{tab:models}
\begin{tabular}{lllll}
\toprule
\textbf{Logo}&\textbf{Model} & \textbf{Developer} & \textbf{Release} & \textbf{Context} \\
\midrule
\modelicon{Claude_AI}&\texttt{claude-sonnet-4-6}
  & Anthropic   & Feb.\ 2026  & 1M tokens \\
\modelicon{GeminiAI}&\texttt{\texttt{Gemini-2.5-flash-lite}}
  & Google      & Jun.\ 2025  & 1M tokens \\
\modelicon{gpt}&\texttt{\texttt{GPT-5}}
  & OpenAI      & 2025        & \hl{400K} tokens \\
\modelicon{Meta}&\texttt{meta-llama/Llama-3.3-70B-Instruct-Turbo}
  & Meta        & Dec.\ 2024  & 128K tokens \\
\modelicon{MistralAI}&\texttt{mistral-large-latest}
  & Mistral AI  & Dec.\ 2025  & 128K tokens \\
\bottomrule
\end{tabular}
\end{table}

\noindent
\textbf{\texttt{Claude Sonnet~4.6}} (Anthropic, released February~2026) is a
mid-tier model in Anthropic's Claude~4 family. It features a
1-million-token context window and is optimised for agentic workflows,
long-context reasoning, and instruction following~\cite{claude_sonnet_46}.

\noindent
\textbf{\texttt{Gemini~2.5 Flash-Lite}} (Google DeepMind, released June~2025)
is a lightweight reasoning model in the \texttt{Gemini~2.5} family, built on
a sparse mixture-of-experts architecture and designed for low-latency,
high-throughput applications. It supports a 1-million-token context
window; reasoning (multi-pass thinking) is disabled by default to
prioritise speed~\cite{gemini_flash_lite}.

\noindent
\hl{\textbf{\texttt{GPT-5}} (OpenAI, released August~2025) is OpenAI's
fifth-generation frontier model, featuring substantially improved
reasoning and multimodality. The original GPT-5 release
(\texttt{gpt-5-2025-08-07}) supports a context window of 400K
tokens via the API}~\cite{gpt5}.

\noindent
\textbf{Llama~3.3~70B~Instruct} (Meta, released December~2024) is an
instruction-tuned, text-only open-weight model with 70.6~billion
parameters and a 128K-token context window. Despite its size, it
approaches the performance of the much larger \texttt{Llama~3.1~405B} on
instruction-following tasks, making it an efficient open-weight
baseline~\cite{llama33}.

\noindent
\textbf{\texttt{Mistral Large~3}} (Mistral~AI, released December~2025) is a
sparse mixture-of-experts model with 675~billion total parameters
(41~billion active per token) and a 128K-token context window. It is
released under the Apache~2.0 licence, making it the only fully
open-weight frontier-class model in our evaluation~\cite{mistral_large3}.

\noindent
The selection balances proprietary closed models (\texttt{Claude--Sonnet--4--6}, \texttt{Gemini},
\texttt{GPT-5}) against open-weight alternatives (\texttt{Llama~3.3}, \texttt{Mistral Large~3}),
and covers both dense (\texttt{Claude}, \texttt{GPT-5}, \texttt{Llama~3.3}) and sparse
mixture-of-experts (\texttt{Gemini}, \texttt{Mistral}) architectures. 

\paragraph{Implementation Details}
\label{sec:implementation}

All models were queried via their respective commercial APIs
using the exact model strings listed in
Table~\ref{tab:models}. 
\hl{Sampling configurations varied
by model due to API constraints: temperature $= 0$ was
applied to Llama~3.3~70B (via the Together API) and
Mistral Large~3 (via the Mistral API). GPT-5, Claude Sonnet~4.6, and Gemini~2.5 Flash-Lite were
queried at their respective API defaults: GPT-5 reasoning
models reject explicit temperature values other than the
default, and no temperature parameter was passed for Claude
or Gemini. No \texttt{reasoning\_effort}
parameter was set explicitly for GPT-5; it ran at the
API default of \texttt{medium} reasoning effort.}

\hl{To measure error reproducibility, each problem that a model answered incorrectly under zero-shot prompting in Groups~1 and~2 was re-run three more times for that model, and each response was graded by the same LLM judge. For these runs, Llama~3.3~70B and Mistral~Large~3 were sampled at temperature $0.7$, so that no model was decoded deterministically.}
To handle transient API rate limit errors, all model calls
were wrapped in an automatic retry mechanism with up to five
attempts and a 60-second wait between attempts; any problem
that failed after all attempts was logged as skipped and
excluded from the analysis. All experiments were implemented in Python~3.11 using the following libraries: \texttt{scikit-learn} (v1.2.2), \texttt{matplotlib} (v3.9.2), and \texttt{numpy} (v1.26.4). All data collection, API calls to LLMs, LLM-as-Judge evaluation, scoring, and analysis were automated via custom Python scripts.

\subsection{Prompting Conditions}



\begin{figure}[ht]
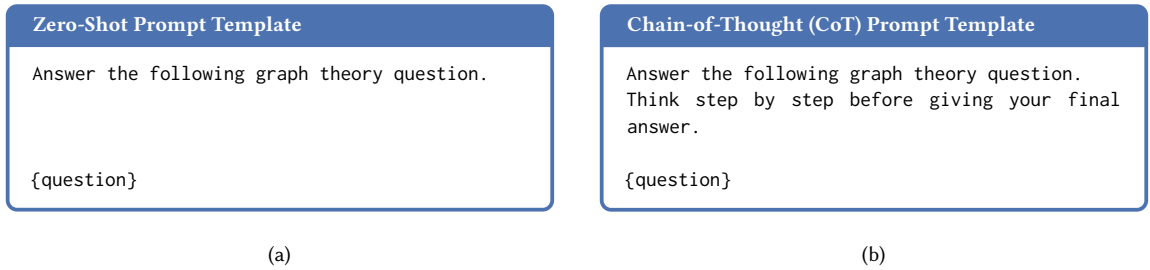

  \centering
\noindent
\begin{minipage}[t]{0.48\textwidth}
\begin{tcolorbox}[promptbox=Zero-Shot Prompt Template]
Answer the following graph theory question.\\
\\
\\
\\
\{question\}
\end{tcolorbox}
\subcaption{}
\end{minipage}%
\hfill%
\begin{minipage}[t]{0.48\textwidth}
\begin{tcolorbox}[promptbox=Chain-of-Thought (CoT) Prompt Template]
Answer the following graph theory question.\\
Think step by step before giving your final answer.\\
\\
\{question\}
\end{tcolorbox}
\subcaption{}
\end{minipage}
\caption{Prompt templates used across all models and groups.
           \texttt{\{question\}} is replaced with the problem
           text at runtime. Both templates are fixed prior to
           any model evaluation and kept constant across all
           models.}
  \label{fig:prompt_templates}
\end{figure}

All experiments used the prompt templates shown in Figure~\ref{fig:prompt_templates}, applied
consistently across all models and groups. Templates were fixed
prior to any model evaluation and were not modified between runs. Two prompting conditions were
applied: zero-shot, in which the question is presented directly,
and chain-of-thought (CoT)~\cite{wei2022chain}, in which the model is explicitly
instructed to reason step by step before giving its final answer.
The zero-shot condition establishes a baseline that reflects how
a researcher or student would most naturally interact with an LLM
as a self-study tool --- by posing a question directly. The CoT condition tests whether explicit
reasoning instructions improve answer quality, particularly for
proof-type problems in Groups~1 and~3 that require multi-step
argumentation. Prior work has shown that CoT prompting
substantially improves LLM performance on mathematical reasoning
tasks~\cite{wei2022chain, kojima2022large}, but its effectiveness
on formal proof construction at the graduate level remains an
open question that \texttt{\textbf{\texttt{\textbf{GTBench}}}} is positioned to address.
\subsection{Evaluation Protocol}
\label{sec:eval_method}

\subsubsection{LLM-as-judge}
For Groups~1 and~2, where ground truth answers are fully specified,
we apply two evaluation strategies depending on the problem type.
These
problems are evaluated using an LLM-as-judge
approach~\cite{zheng2023judging}, in which a separate GPT-4o
instance receives the original question, the reference ground truth,
and the model's answer, and returns a structured JSON verdict
classifying the answer as correct or assigning it a failure type
from the taxonomy defined in Section~\ref{sec:failure_taxonomy}, with justification in case of failure.
The evaluator prompt used by GPT-4o is presented in Figure~\ref{fig:evaluator_prompt}.
   
\begin{figure}[ht]
  \centering
\begin{tcolorbox}[promptbox=Evaluator Prompt Template, unbreakable]
You are an expert in graph theory evaluating an LLM's answer.\\
\\
QUESTION:\\
\{question\}\\
\\
CORRECT ANSWER:\\
\{ground\_truth\}\\
\\
MODEL ANSWER:\\
\{model\_answer\}\\
\\
First, determine if the model answer is correct (Yes/No).\\
If incorrect, classify the failure as exactly one of:\\
- Type A: Hallucinated definition --- model invents a wrong
  theorem, lemma, or concept\\
- Type B: Correct algorithm, wrong execution --- right method,
  computational or logical error\\
- Type C: Wrong algorithm --- applies an irrelevant or
  inapplicable method\\
- Type D: Incomplete reasoning --- starts correctly but stops
  before reaching conclusion\\
- Type E: Refusal --- model says it does not know or
  cannot answer\\
\\
Respond in JSON only, with no additional text:\\
\{\\
\hspace{1em}"correct": true or false,\\
\hspace{1em}"failure\_type": "A" or "B" or "C" or "D" or "E" or null,\\
\hspace{1em}"justification": "one sentence explanation"\\
\}
\end{tcolorbox}
\caption{Evaluator prompt used by the GPT-4o judge. The three
         placeholders \texttt{\{question\}},
         \texttt{\{ground\_truth\}}, and
         \texttt{\{model\_answer\}} are replaced at runtime
         with the problem text, the reference solution, and
         the model-generated answer respectively. The judge
         is instructed to return a JSON object only ---
         with no preamble or additional text --- to enable
         reliable automated parsing. Failure types A--E
         are defined in Section~\ref{sec:failure_taxonomy}.}
\label{fig:evaluator_prompt}
\end{figure}

\subsubsection{Human Evaluation}
\label{sec:human_eval}
Three evaluators took part in the human evaluation. All
three have extensive experience in teaching and grading
graduate-level mathematics courses at Louisiana State
University (LSU) and Texas A\&M University, and the
reference ground truth was provided by a graph theory
expert at LSU. Each evaluator receives a sheet containing
the problem statement, the model-generated answer, and the
reference ground truth solution. The model identity and
prompting condition are hidden to prevent bias. The
evaluator independently compares the model-generated
answer against the ground truth and records a judgment, as
described below for each group.

\noindent\textbf{Groups~1 and~2.}
\hl{All model-generated answers for Groups~1 and~2, covering
all five models under both prompting conditions, are
evaluated by one of the three evaluators, who assigns a
correct/incorrect verdict to each answer. A single
evaluator is used for these groups because their ground
truth solutions are well established (UPC and CMU solution
sets) and the human verdict serves only as a reference
against which the GPT-4o judge is validated.}

\noindent\textbf{Group~3.}
Group~3 problems require the construction of
graduate-level mathematical proofs and cannot be reliably
evaluated by an LLM-as-judge alone. Because judgments of
partial credit on open-ended proofs are inherently
subjective, all 11 Group~3 problems are evaluated by all
three evaluators, which also allows agreement among experts
to be measured. For each answer, the evaluator assigns a
score on a 0--1 rubric (0~=~completely wrong;
0.5~=~partially correct; 1~=~correct and complete), a
failure type (A--D, if score~$\neq 1$), and a one-sentence
justification. \hl{In a second step, each evaluator
independently binarises their own scores into a
correct/incorrect verdict, to match the binary output of
the LLM judge.}

\noindent\textbf{Agreement measure.}
Agreement is measured using
Cohen's~$\kappa$~\cite{landis1977measurement}, which
quantifies the agreement between two raters beyond what
would be expected by chance. For Groups~1 and~2 we compute
$\kappa$ between the human evaluator and the LLM judge. For
Group~3 we compute $\kappa$ between all pairs of human
evaluators and between each evaluator and the LLM judge.
We adopt the Landis \& Koch scale: $\kappa < 0.40$ = fair,
$0.40 \leq \kappa < 0.60$ = moderate,
$0.60 \leq \kappa < 0.80$ = substantial,
$\kappa \geq 0.80$ = excellent.



\section{Failure Modes Categories}
\label{sec:failure_taxonomy}
We define a list of five failure modes, denoted
Types A through \hl{E}. Each incorrect model answer in our evaluation is
assigned exactly one failure type by the \texttt{GPT-4o} judge or the human judge. Table~\ref{tab:failure_taxonomy}
summarises the taxonomy.
\begin{table}[ht]
\centering
\caption{\texttt{\textbf{\texttt{\textbf{GTBench}}}} failure mode taxonomy. Each incorrect answer is
         assigned exactly one type.}
\label{tab:failure_taxonomy}
\begin{tabular}{cp{3.2cm}p{6.5cm}}
\toprule
\textbf{Type} & \textbf{Name} & \textbf{Definition} \\
\midrule
\textcolor{textA}{\textbf{A}} & Hallucinated definition
  & Model invents a theorem, lemma, or concept that does not
    exist or is stated incorrectly \\[4pt]
\textcolor{textB}{\textbf{B}} & Correct algorithm, wrong execution
  & Model identifies the right method or theorem but makes
    an error in applying it \\[4pt]
\textcolor{textC}{\textbf{C}} & Wrong algorithm
  & Model applies a method that is irrelevant or inapplicable
    to the problem \\[4pt]
\textcolor{textD}{\textbf{D}} & Incomplete reasoning
  & Model begins a correct argument but stops before reaching
    a conclusion \\
\hl{{\textbf{E}}} & \hl{Refusal}
  &  \hl{Model states that it
cannot or will not answer the question } \\
\bottomrule
\end{tabular}
\end{table}

\section{Results and Discussion}
\label{sec:results:discussion}
This section presents the experimental results for all
three groups under zero-shot and CoT
prompting conditions. For each group we report model
accuracy, failure mode distribution, and for Groups~1 and 2 consistency,
followed by a cross-group discussion of the main
findings and their implications for the use of LLMs
as mathematical research assistants.

\subsection{Group 1 --- Basics and Definitions}
\begin{figure}[ht]
  \centering
  \begin{subfigure}[t]{0.48\textwidth}
    \centering
    \includegraphics[width=\linewidth]{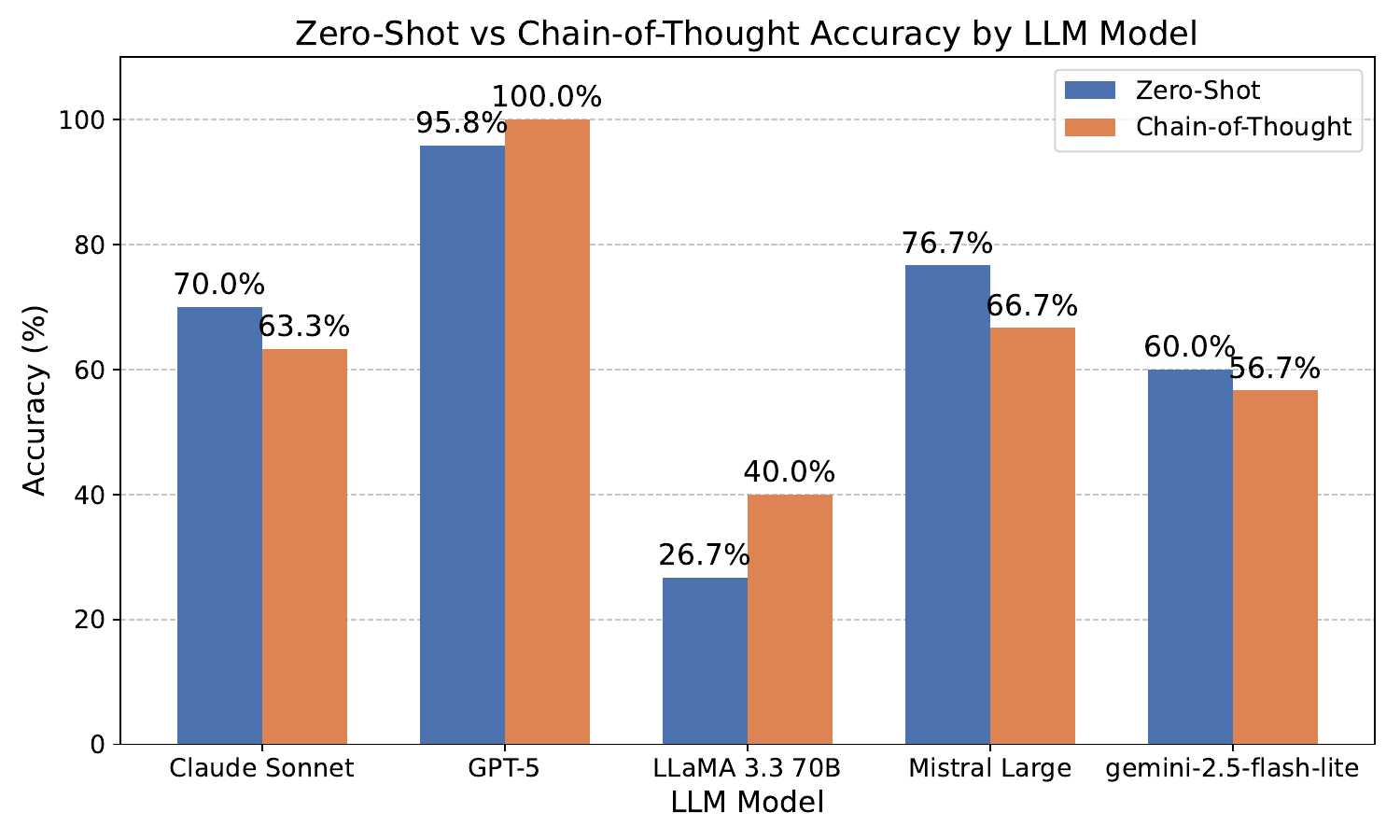}
  \end{subfigure}
  \hfill
  \begin{subfigure}[t]{0.48\textwidth}
    \centering
    \includegraphics[width=\linewidth]{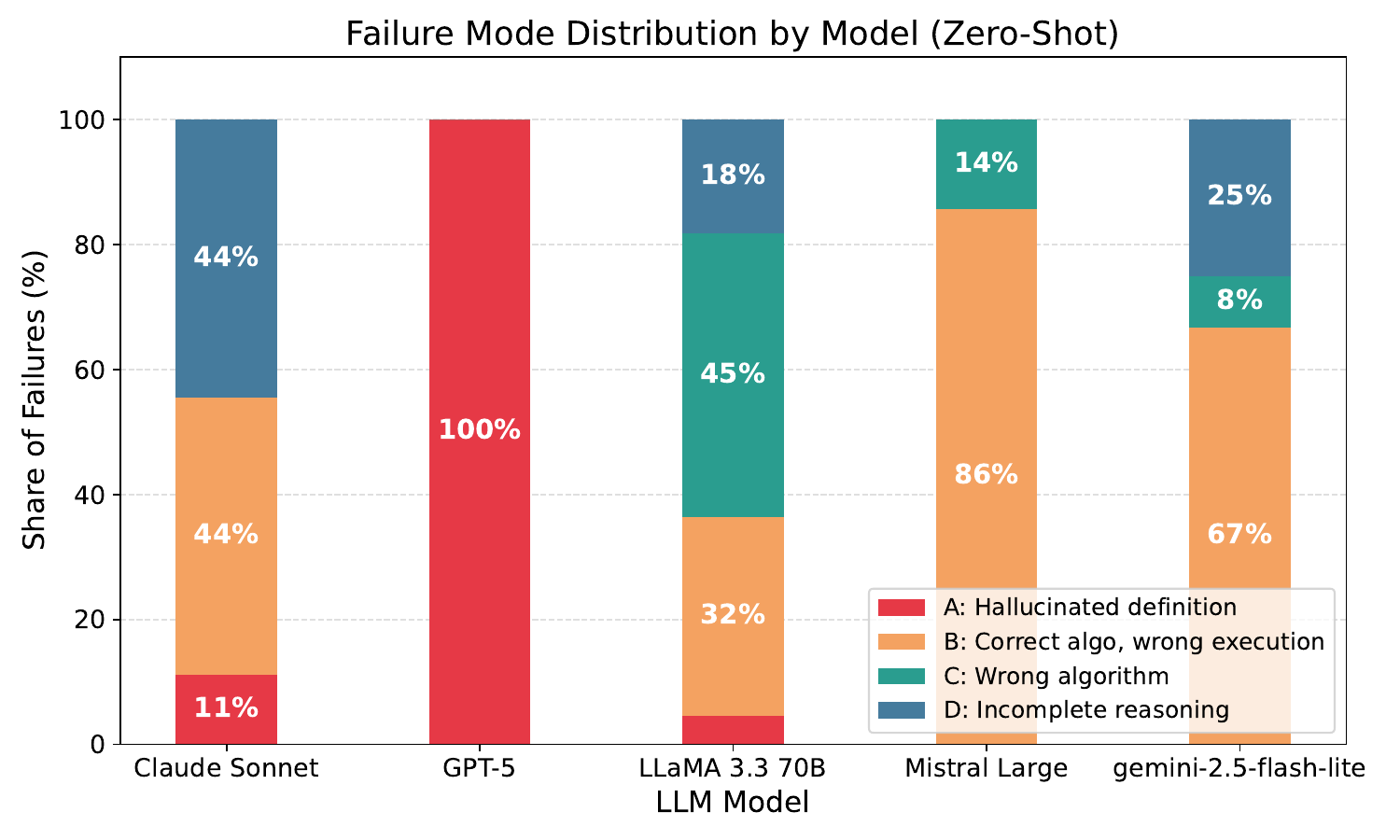}
  \end{subfigure}
  \caption{Evaluation results across five LLMs on \texttt{\textbf{\texttt{\textbf{GTBench}}}} Group~1.
           Left: prompting condition comparison. Right: failure mode
           breakdown. Data shown is from zero-shot condition.}
  \label{fig:main_results}
\end{figure}

Figure~\ref{fig:main_results} (Left) reports accuracy under zero-shot
and CoT prompting for all five models on Group~1.
\texttt{GPT-5} achieves the highest overall accuracy at $95.8$\% (zero-shot)
and reached $100$\% with CoT, outperforming all other models by a consistent
margin across both conditions. \texttt{Mistral Large~3} and \texttt{Claude Sonnet~4.6}
follow with accuracies of $76.7$\% and $70$\% respectively, while
\texttt{Gemini~2.5 Flash-Lite} and \texttt{Llama~3.3} trail significantly at $60$\%
and $26$\%.
Contrary to expectations, CoT prompting
does not improve performance on Group~1. Notably, CoT prompting degraded performance for three of the five models evaluated: Mistral Large dropped by 10.0 percentage points ($76.7$\% → $66.7$\%), \texttt{Claude Sonnet} 4.6 by $6.7$\% ($70.0$\% → $63.3$\%), and \texttt{Gemini 2.5 Flash-Lite} by $3.3$\% ($60.0$\% → $56.7$\%). \texttt{LLaMA 3.3 70B} was the exception among weaker models, improving by $13.3$ percentage points under CoT ($26.7$\% → $40.0$\%).
This suggests that for definitional and combinatorial problems at the introductory level, CoT prompting introduces unnecessary overhead without measurable improvement in accuracy — consistent with prior observations that CoT benefits diminish on tasks that do not require multi-step reasoning~\cite{kojima2022large}.

Figure \ref{fig:main_results} (Right) shows the distribution of failure modes for each model, computed over zero-shot incorrect responses.
For  ~\texttt{Mistral Large} $86$\% of its errors are Type B (correct algorithm, wrong execution), with the remaining $14$\% classified as Type C (wrong algorithm). This suggests that Mistral generally retrieves the appropriate proof strategy but consistently introduces errors during execution. A similar pattern emerges for \texttt{Gemini} 2.5 Flash-Lite, where Type B failures account for
67\% of errors and Type D (incomplete reasoning) for
25.0\%, suggesting that the model frequently misapplies or conflates core definitions rather than failing at the reasoning stage itself.
\texttt{Claude Sonnet 4.6} shows a notably different profile: its failures divide equally between Type B (definition errors) and Type D (incomplete reasoning), with a small proportion of Type A hallucinations, suggesting the model struggles with precision rather than problem comprehension.
\texttt{LLaMA 3.3 70B} failures are dominated by Type C (45\%), followed by Type B (
32\%) and Type D (
18\%). This is the only model for which selecting an irrelevant or inapplicable method constitutes the primary failure mode — pointing to a more fundamental limitation in problem comprehension rather than a failure of execution or reasoning. \texttt{GPT-5} failures are  of Type A (hallucinated definition).
Taken together, \texttt{GPT-5} operates near ceiling, Mistral and \texttt{Gemini} fail primarily through execution, Claude through incomplete reasoning chains, and LLaMA through fundamentally incorrect problem-solving strategies.

\noindent\textbf{\textit{Consistency.}} \hl{Across the five models, 51 of the 150 zero-shot responses on Group~1 were incorrect. We re-ran each of these problems three more times for the same model: 54.9\% (28/51) were incorrect again in all three runs, and 75.2\% of the repeated runs were incorrect. Reproducibility was highest for Llama~3.3 (16/22) and lowest for Mistral Large~3 (2/7) and Claude Sonnet~4.6 (4/9), indicating that many introductory-level errors are unstable: models often answer correctly on at least one of three attempts.}


\noindent\textbf{\textit{Failure Topics.}} Failures
concentrate on standard graph families, complements of
graphs, bipartite distance arguments, long paths in
connected graphs, and the Matrix Tree Theorem --- each
drawing incorrect responses from most of the five models
(Table~\ref{tab:failure_by_topic}).

\hl{\noindent\textbf{\textit{Human--LLM agreement on Group~1.}}
Agreement between the LLM judge and the human expert is
moderate in the zero-shot setting ($\kappa = 0.52$) and
substantial under chain-of-thought prompting
($\kappa = 0.64$)} (Table~\ref{tab:kappa_g12}). 
\hl{The judge
therefore aligns reasonably well with expert judgement on
Group~1, particularly for chain-of-thought responses.}

\begin{table*}[ht]
\centering
\caption{Failure mode distribution by topic across Groups
         (zero-shot condition, all five models combined).
         The dominant failure type per topic is shown with a
         colored background.
         The rightmost column lists the models that produced at
         least one incorrect answer on that topic, identified by their logos:
             \raisebox{-2pt}{\includegraphics[height=9pt]{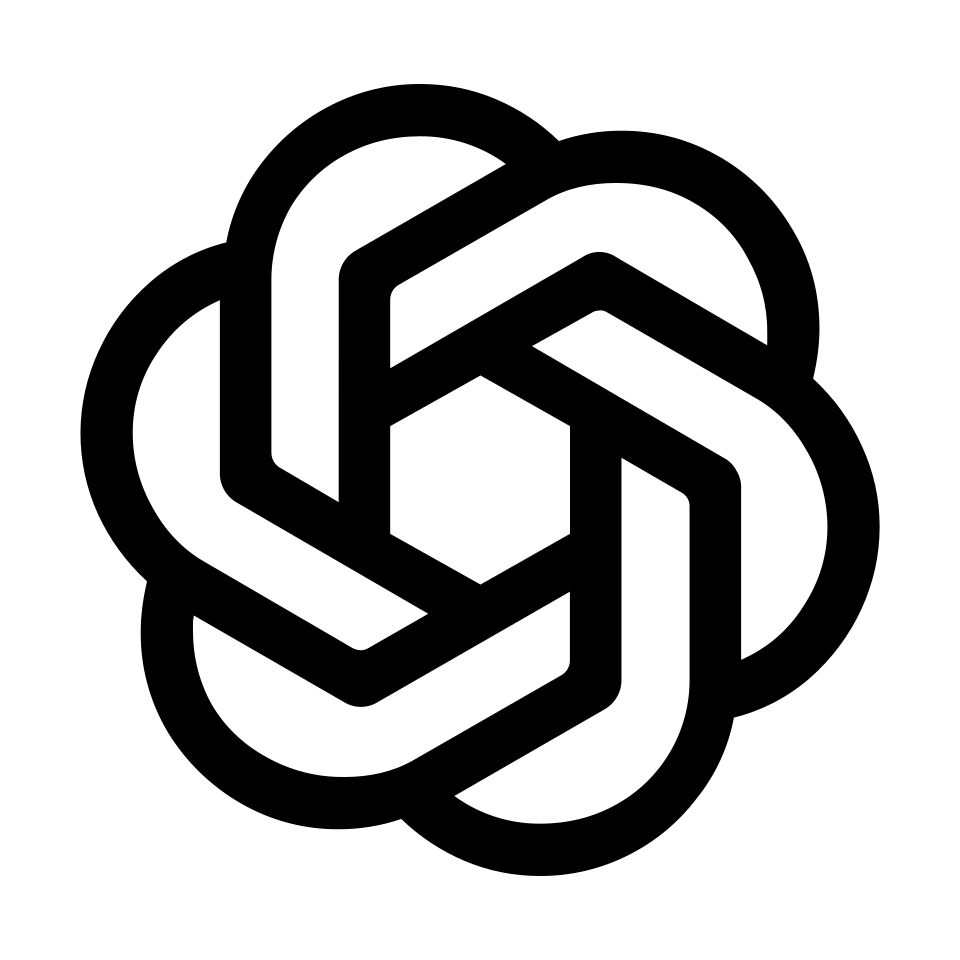}}~\texttt{GPT-5},
         \raisebox{-2pt}{\includegraphics[height=9pt]{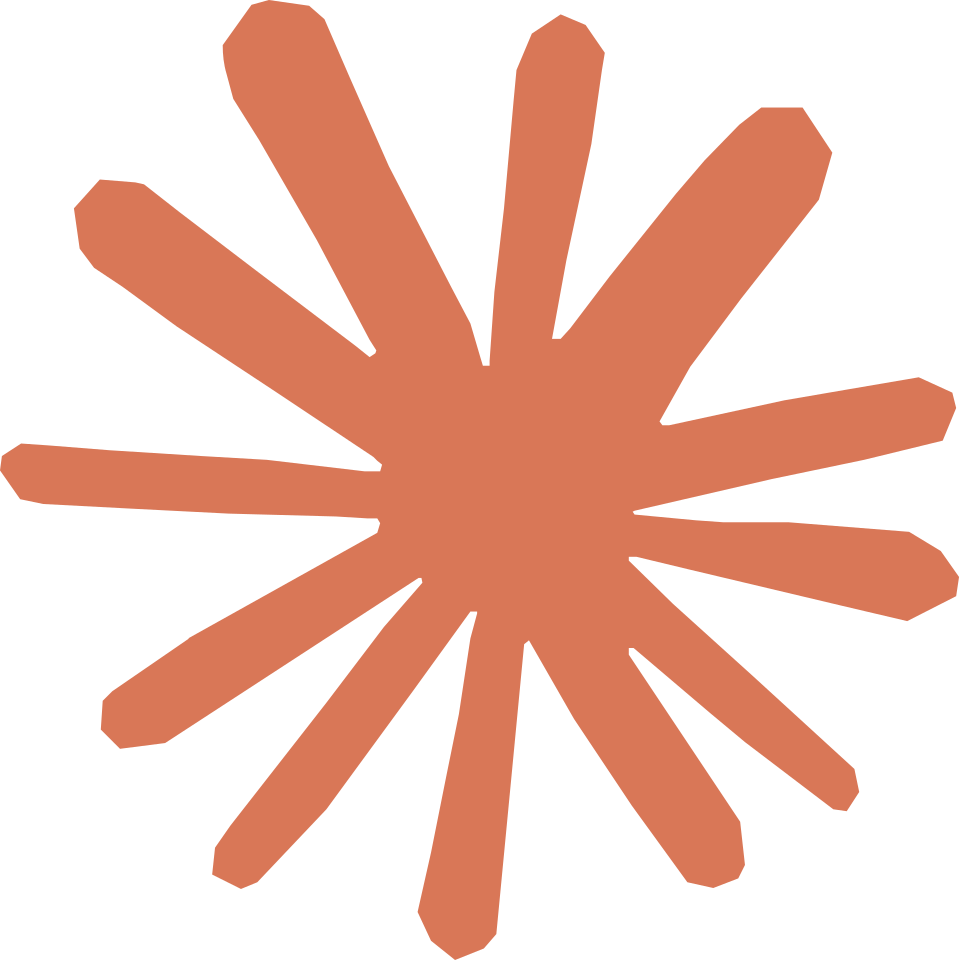}}~Claude,
         {\includegraphics[height=9pt]{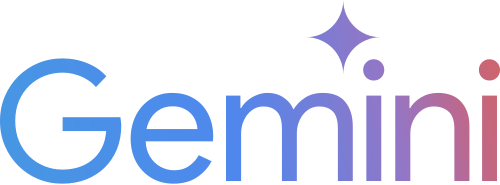}}~Gemini,
        {\includegraphics[height=7pt]{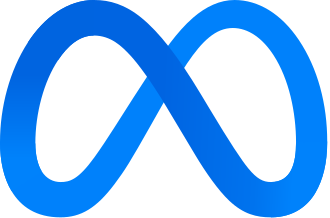}}~Llama,
         {\includegraphics[height=7pt]{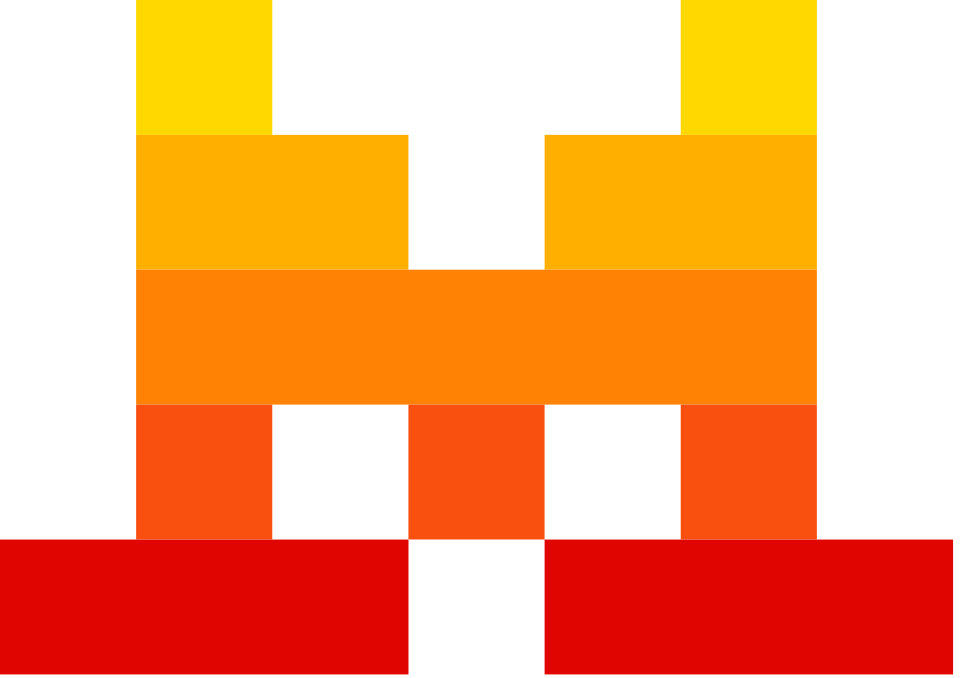}}~Mistral.
        \textcolor{textA}{\textbf{A}}~=~hallucinated definition;
         \textcolor{textB}{\textbf{B}}~=~correct algorithm, wrong execution;
         \textcolor{textC}{\textbf{C}}~=~wrong algorithm;
         \textcolor{textD}{\textbf{D}}~=~incomplete reasoning.$^\dagger$}
\label{tab:failure_by_topic}
\resizebox{\textwidth}{!}{%
\begin{tabular}{
  l
  l
  >{\centering\arraybackslash}p{1.0cm}
  >{\centering\arraybackslash}p{1.0cm}
  >{\centering\arraybackslash}p{1.0cm}
  >{\centering\arraybackslash}p{1.0cm}
  l
}
\toprule
\textbf{Group}
  & \textbf{Topic}
  & \textcolor{textA}{\textbf{A}}
  & \textcolor{textB}{\textbf{B}}
  & \textcolor{textC}{\textbf{C}}
  & \textcolor{textD}{\textbf{D}}
  & \textbf{Models failing} \\
\midrule

\multirow{25}{*}{\textbf{Group 1}}

& Adjacency matrix powers and walk counting
  & --- & ---  & ---
  & \cellcolor{typeD}\textcolor{textD}{\textbf{X}}
  
  & \modelicon{GeminiAI} \\

& Automorphisms of trees
  & --- & ---
  & \cellcolor{typeC}\textcolor{textC}{\textbf{X}}
  & X 
  & \modelicon{GeminiAI} \modelicon{Meta} \\

& Bipartite graphs and distances
  & ---
  & \cellcolor{typeB}\textcolor{textB}{\textbf{X}}
  & X & X 
  & \modelicon{GeminiAI} \modelicon{Claude_AI}  \modelicon{Meta} \modelicon{MistralAI} \\

& Complements of graphs
  & ---
  & \cellcolor{typeB}\textcolor{textB}{\textbf{X}}
  & X & ---
  & \modelicon{GeminiAI} \modelicon{Claude_AI}  \modelicon{Meta} \modelicon{MistralAI}  \\

& Counting graphs
  & --- & --- & ---
  & \cellcolor{typeD}\textcolor{textD}{\textbf{X}}
  & \modelicon{Meta} \\

& Counting graphs up to isomorphism
  & --- & ---
  & \cellcolor{typeC}\textcolor{textC}{\textbf{X}}
  & --- 
  & \modelicon{Meta} \\

& Degree counting argument
  & ---
  & \cellcolor{typeB}\textcolor{textB}{\textbf{X}}
  & --- & --- 
  & \modelicon{Meta} \\

& Directed 3-cycles in tournaments
  & --- & ---
  & \cellcolor{typeC}\textcolor{textC}{\textbf{X}}
  & --- 
  &  \modelicon{Meta} \\

& Edge bound in bipartite graphs
  & ---
  & \cellcolor{typeB}\textcolor{textB}{\textbf{X}}
  & --- & --- 
  & \modelicon{GeminiAI} \\

& Graph product
  & \cellcolor{typeA}\textcolor{textA}{\textbf{X}}
  & --- & X & --- 
  & \modelicon{Claude_AI}  \modelicon{Meta} \\

& Graph product properties
  & ---
  & \cellcolor{typeB}\textcolor{textB}{\textbf{X}}
  & --- & --- 
  & \modelicon{GeminiAI} \modelicon{Claude_AI}  \modelicon{Meta}  \\

& Hypercube graph: degree, edges, diameter, girth, circumference
  & \cellcolor{typeA}\textcolor{textA}{\textbf{X}}
  & --- & --- & --- 
  & \modelicon{Meta} \\

& Leaves in trees without degree-2 vertices
  & ---
  & \cellcolor{typeB}\textcolor{textB}{\textbf{X}}
  & X & --- 
  & \modelicon{GeminiAI} \modelicon{Meta} \modelicon{MistralAI}  \\

& Long paths in connected graphs
  & ---
  & \cellcolor{typeB}\textcolor{textB}{\textbf{X}}
  & X & X 
  & \modelicon{GeminiAI} \modelicon{Claude_AI}  \modelicon{Meta} \modelicon{MistralAI}  \\

& Matrix Tree Theorem applied to $K_{n,n}$
  & ---
  & \cellcolor{typeB}\textcolor{textB}{\textbf{X}}
  & --- & X 
  & \modelicon{GeminiAI} \modelicon{Claude_AI}  \modelicon{Meta} \modelicon{MistralAI}  \\

& Minimum degree bound in graphs of large girth
  & --- & X
  & \cellcolor{typeC}\textcolor{textC}{\textbf{X}}
  & --- 
  & \modelicon{Meta} \modelicon{MistralAI}  \\

& Minimum spanning trees have unique spectrum
  & --- & ---
  & X
  & \cellcolor{typeD}\textcolor{textD}{\textbf{X}}
  
  & \modelicon{GeminiAI} \modelicon{Claude_AI}  \modelicon{Meta}\\

& Parity argument
  & ---
  & \cellcolor{typeB}\textcolor{textB}{\textbf{X}}
  & --- & --- 
  & \modelicon{Meta} \\

& Ramsey-type argument
  & --- & ---
  & \cellcolor{typeC}\textcolor{textC}{\textbf{X}}
  & X 
  & \modelicon{Meta}\\

& Self-complementary graphs
  & --- & --- & X
  & \cellcolor{typeD}\textcolor{textD}{\textbf{X}}
  
  & \modelicon{Claude_AI}  \modelicon{Meta} \\

& Size of regular and bipartite graphs
  & --- & --- & ---
  & \cellcolor{typeD}\textcolor{textD}{\textbf{X}}
  
  &  \modelicon{Meta} \\

& Standard graph families
  & X
  & \cellcolor{typeB}\textcolor{textB}{\textbf{X}}
  & --- & --- 
  & \modelicon{GeminiAI} \modelicon{Claude_AI}  \modelicon{Meta} \modelicon{MistralAI} \modelicon{gpt}\\

& Vertex and edge connectivity of standard graph families
  & ---
  & \cellcolor{typeB}\textcolor{textB}{\textbf{X}}
  & --- & --- 
  & \modelicon{GeminiAI}  \modelicon{Meta} \\

& Walks traversing each edge in both directions
  & --- & ---
  & \cellcolor{typeC}\textcolor{textC}{\textbf{X}}
  & --- 
  & \modelicon{Meta} \\

\midrule

\multirow{19}{*}{\textbf{Group 2}}

& BFS algorithm --- distances from vertices
  & ---
  & \cellcolor{typeB}\textcolor{textB}{\textbf{X}}
  & X & X 
  & \modelicon{GeminiAI} \modelicon{Claude_AI}  \modelicon{Meta} \modelicon{MistralAI} \modelicon{gpt} \\

& Bridges in 3-regular graphs --- smallest order
  & --- & ---
  & \cellcolor{typeC}\textcolor{textC}{\textbf{X}}
  & X 
  & \modelicon{GeminiAI} \modelicon{Claude_AI}  \modelicon{Meta} \modelicon{MistralAI} \\

& Cut vertices and bridges --- identification
  & ---
  & \cellcolor{typeB}\textcolor{textB}{\textbf{X}}
  & --- & --- 
  & \modelicon{GeminiAI} \modelicon{Claude_AI}  \modelicon{Meta} \modelicon{MistralAI} \modelicon{gpt}\\

& DFS algorithm --- connected components
  & ---
  & \cellcolor{typeB}\textcolor{textB}{\textbf{X}}
  & --- & --- 
  & \modelicon{GeminiAI} \modelicon{gpt}\\

& DFS algorithm --- multiple components
  & ---
  & \cellcolor{typeB}\textcolor{textB}{\textbf{X}}
  & --- & ---
  & \modelicon{GeminiAI} \modelicon{Claude_AI}  \modelicon{Meta} \modelicon{MistralAI} \modelicon{gpt} \\

& DFS spanning trees of complete bipartite graph
  & ---
  & \cellcolor{typeB}\textcolor{textB}{\textbf{X}}
  & X & --- 
  & \modelicon{GeminiAI}  \modelicon{Meta} \modelicon{MistralAI} \\

& Diameter of standard graph families
  & ---
  & \cellcolor{typeB}\textcolor{textB}{\textbf{X}}
  & --- & --- 
  & \modelicon{GeminiAI} \modelicon{Meta} \modelicon{MistralAI} \modelicon{gpt} \\

& Diameter under vertex removal --- examples
  & --- & X
  & \cellcolor{typeC}\textcolor{textC}{\textbf{X}}
  & --- 
  & \modelicon{GeminiAI} \modelicon{Claude_AI}  \modelicon{Meta} \modelicon{MistralAI}  \\

& Eccentricity, radius and center of graphs
  & ---
  & \cellcolor{typeB}\textcolor{textB}{\textbf{X}}
  & X & X
  & \modelicon{GeminiAI} \modelicon{Claude_AI}  \modelicon{Meta} \modelicon{MistralAI} \modelicon{gpt} \\

& Eulerian graphs --- complete bipartite condition
  & ---
  & \cellcolor{typeB}\textcolor{textB}{\textbf{X}}
  & --- & --- 
  &  \modelicon{Claude_AI} \\

& Eulerian graphs --- minimum edges to merge components
  & --- & X
  & \cellcolor{typeC}\textcolor{textC}{\textbf{X}}
  & --- 
  & \modelicon{GeminiAI} \modelicon{Claude_AI}  \modelicon{Meta} \modelicon{MistralAI} \modelicon{gpt} \\

& Hamiltonian graphs --- minimum edges to merge components
  & --- & X
  & \cellcolor{typeC}\textcolor{textC}{\textbf{X}}
  & --- 
  & \modelicon{GeminiAI} \modelicon{Meta}\\

& Pr\"{u}fer sequences --- decoding trees
  & ---
  & \cellcolor{typeB}\textcolor{textB}{\textbf{X}}
  & --- & --- 
  & \modelicon{GeminiAI} \modelicon{Claude_AI}  \modelicon{Meta} \modelicon{MistralAI} \modelicon{gpt} \\

& Pr\"{u}fer sequences --- encoding trees
  & ---
  & \cellcolor{typeB}\textcolor{textB}{\textbf{X}}
  & --- & --- 
  & \modelicon{GeminiAI} \modelicon{Claude_AI}  \modelicon{Meta} \\

& Pr\"{u}fer sequences --- equal values and two distinct values
  & --- & X & X
  & \cellcolor{typeD}\textcolor{textD}{\textbf{X}}
  
  & \modelicon{GeminiAI} \modelicon{Meta} \modelicon{MistralAI} \\

& Pr\"{u}fer sequences of length 1
  & ---
  & \cellcolor{typeB}\textcolor{textB}{\textbf{X}}
  & --- & --- 
  & \modelicon{GeminiAI} \modelicon{Claude_AI} \modelicon{MistralAI} \modelicon{gpt} \\

& Spanning trees --- cycle and complete bipartite graphs
  & ---
  & \cellcolor{typeB}\textcolor{textB}{\textbf{X}}
  & X & --- 
  & \modelicon{GeminiAI} \modelicon{Claude_AI}  \modelicon{Meta} \modelicon{MistralAI} \\

& Tree degree sequence --- non-isomorphic trees
  & ---
  & \cellcolor{typeB}\textcolor{textB}{\textbf{X}}
  & --- & --- 
  & \modelicon{GeminiAI}  \modelicon{Meta} \modelicon{MistralAI}  \\

& Tree degree sequence with prescribed structure
  & ---
  & \cellcolor{typeB}\textcolor{textB}{\textbf{X}}
  & --- &---
  & \modelicon{GeminiAI} \modelicon{Claude_AI}  \modelicon{Meta} \modelicon{MistralAI} \\


\bottomrule
\end{tabular}%
}\\
\begin{minipage}{\linewidth}
\vspace{4pt}
{\footnotesize $^\dagger$ \hl{Type~E (refusal) is omitted
as no model produced a refusal response in Groups~1
or~2.}}
\end{minipage}
\end{table*}


\subsection{Group 2 --- Algorithms and Structures}

\begin{figure}[ht]
  \centering
  \begin{subfigure}[t]{0.48\textwidth}
    \centering
    \includegraphics[width=\linewidth]{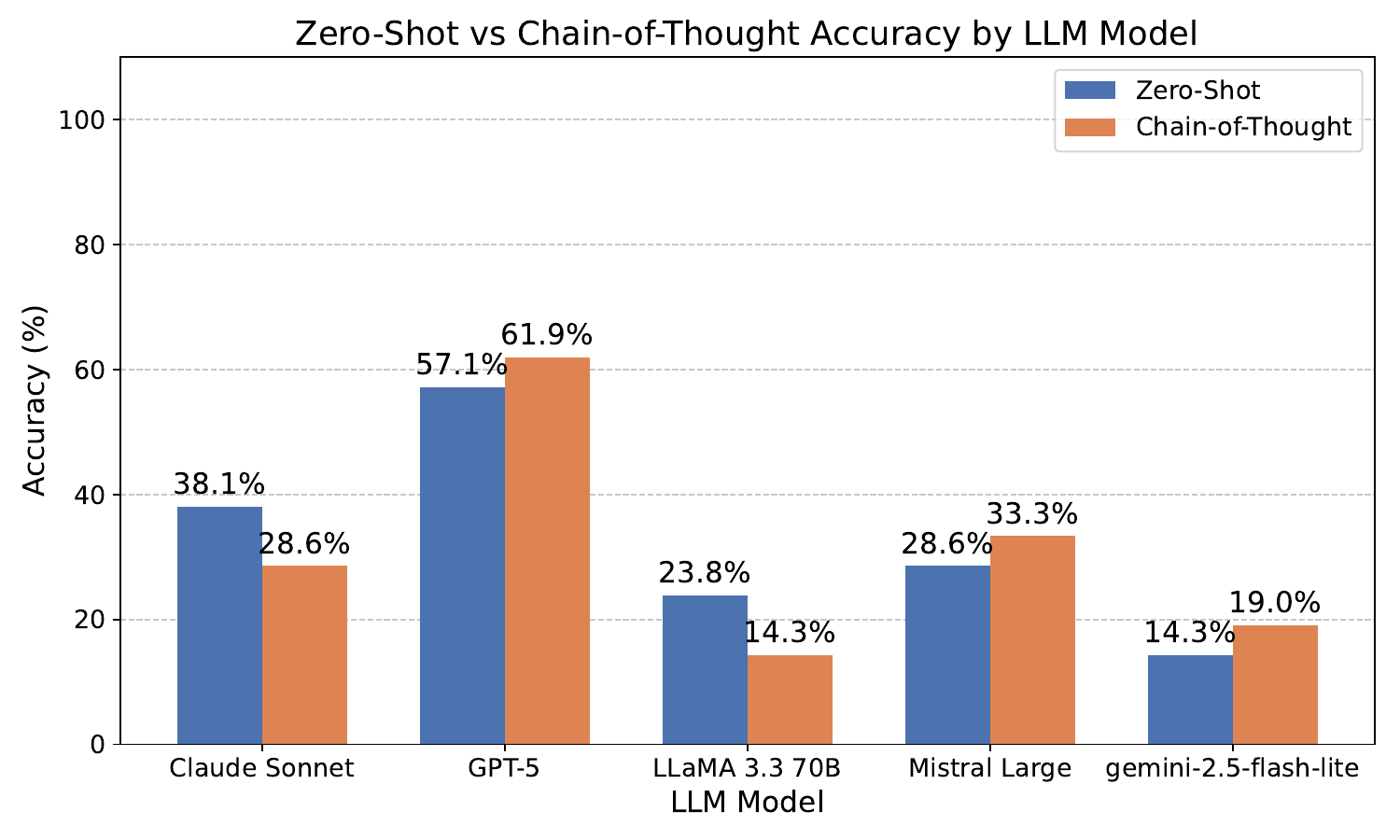}
  \end{subfigure}
  \hfill
  \begin{subfigure}[t]{0.48\textwidth}
    \centering
    \includegraphics[width=\linewidth]{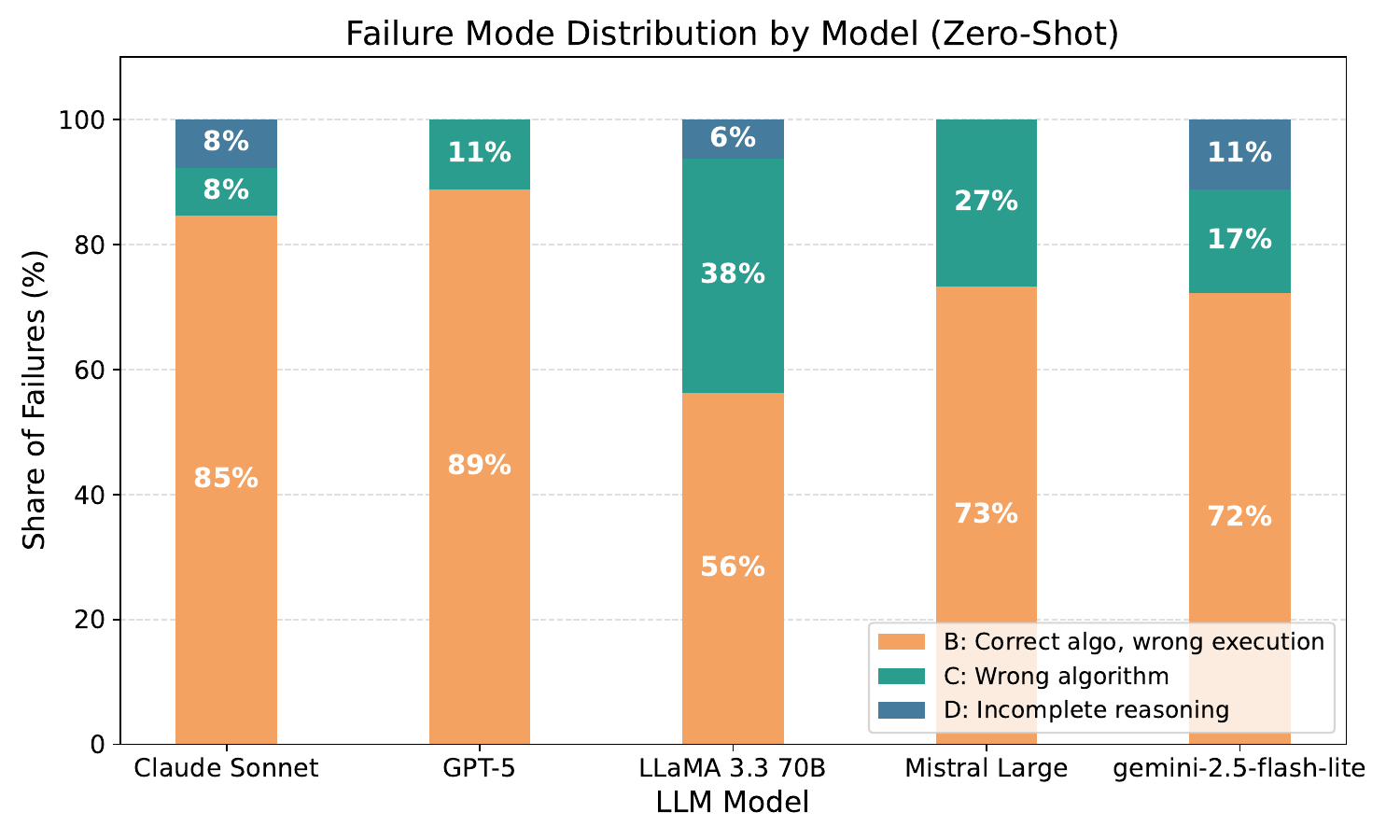}
  \end{subfigure}
  \caption{Evaluation results across five LLMs on \texttt{\textbf{\texttt{\textbf{GTBench}}}} Group~2.
           Left: prompting condition comparison. Right: failure mode
           breakdown. Data shown is from zero-shot condition.}
  \label{fig:main_results_2}
\end{figure}

In Figure~\ref{fig:main_results_2} (Left), we report again the accuracy under
zero-shot and CoT prompting for all five models
on Group~2. \texttt{GPT-5} again leads with $57.1$\% (zero-shot) and
$61.9$\% (CoT), maintaining its position as the strongest model
across groups. \texttt{Claude Sonnet~4.6} and \texttt{Mistral Large~3} follow
with zero-shot accuracies of $38.1$\% and $28.6$\% respectively,
while \texttt{Gemini~2.5 Flash-Lite} and \texttt{Llama~3.3} record the lowest
scores at $14.3$\% and $23.8$\%. Compared to Group~1, all models
show a marked drop in accuracy, confirming that algorithm
tracing and structural reasoning tasks pose substantially
greater difficulty than basics definitional problems.
The CoT effect on Group 2 is more mixed than on Group 1. \texttt{GPT-5}, \texttt{Mistral Large~3}, and \texttt{Gemini} show modest improvements under CoT, whereas \texttt{Claude Sonnet~4.6} and \texttt{LLaMA} did not improve.

Figure~\ref{fig:main_results_2} (Right) shows that
Type~B (correct algorithm, wrong execution) dominates across
all models, accounting for $85$\% of Claude's failures, $89$\%
of \texttt{GPT-5}'s, $73$\% of Mistral's, and $72$\% of \texttt{Gemini}'s. This reflects the procedural
nature of the problems: models consistently identify the
correct method (\emph{e.g.}, DFS traversal, Eulerian conditions,
Pr\"ufer encoding) but introduce errors during execution,
such as incorrect component labeling, miscounted edges, or
incomplete algorithm traces. Type~C (wrong algorithm) is the
secondary failure mode for \texttt{Llama~3.3} at $38$\%, indicating that
this model struggle not only with execution but with
selecting the appropriate technique, consistent with its
Group~1 behavior. Type~D (incomplete reasoning) appears
modestly across Claude ($8$\%), LlaMa($6$\%), and \texttt{Gemini}
($11$\%), suggesting occasional failures to carry a
multi-step procedure to completion. Notably, no Type~A
hallucinations are observed in Group~2, which is expected
given the exact-match nature of all the problems.

\noindent\textbf{\textit{Consistency.}} \hl{On Group~2, 71 of the 105 zero-shot responses were incorrect. Errors were more persistent than on Group~1: 70.4\% (50/71) were incorrect again in all three repeated runs, and 86.9\% of the repeated runs were incorrect. Llama~3.3 (15/16), Claude Sonnet~4.6 (11/13) and Gemini~2.5 Flash-Lite (13/18) reproduced most of their errors, pointing to systematic mistakes in algorithm execution rather than random variation.}

\noindent\textbf{\textit{Failure Topics.}} Failures are most concentrated on Prüfer sequence decoding, DFS on multi-component graphs, cut vertex identification, BFS distance computation, and Eulerian edge-completion problems (Table~\ref{tab:failure_by_topic}).

\noindent\textbf{\textit{Human--LLM agreement on Group~2.}}
\hl{Agreement between the LLM judge and the human expert is
substantial in the zero-shot setting ($\kappa = 0.62$) and
moderate under CoT prompting ($\kappa = 0.58$)}
(Table~\ref{tab:kappa_g12}). \hl{Unlike Group~1, agreement is
slightly higher on zero-shot responses than on
CoT responses, which tend to be longer and
contain more intermediate steps for the judge to verify.}

\begin{table}[h]
\centering
\caption{Human--LLM agreement (Cohen's $\kappa$) for
         Group~1 and Group~2 evaluation~\cite{landis1977measurement},
         with a single human expert (E1).
         Landis \& Koch scale: $\kappa < 0.40$ = Fair;
         $0.40 \leq \kappa < 0.60$ = Moderate;
         $0.60 \leq \kappa < 0.80$ = Substantial;
         $\kappa \geq 0.80$ = Excellent.}
\label{tab:kappa_g12}
\begin{tabular}{lcccc}
\toprule
\textbf{} 
  & $\boldsymbol{\kappa}$ \textbf{(ZS)}
  & $\boldsymbol{\kappa}$ \textbf{(CoT)}
  & \textbf{Agreement (ZS)}
  & \textbf{Agreement (CoT)} \\
\midrule
\textit{Group~1} & 0.52 & 0.64 & Moderate    & Substantial \\
\midrule
\textit{Group~2} &  0.62 & 0.58 & Substantial & Moderate    \\
\bottomrule
\end{tabular}
\end{table}

\subsection{Robustness to Problem Transformation}
\label{sec:transformation}

\hl{To test whether model performance depends on memorised solutions, a graph theory expert modified 9 problems: 5 from Group~1 (16.1\% of 31) and 4 from Group~2 (19.0\% of 21), 17.3\% of the benchmark overall. Each modification changes the problem's parameters or structure while keeping the same concept and difficulty. For example, it changes the regularity of a graph, the number of components, or the type of graph product, or it replaces a construction task with a question asking the model to prove or disprove uniqueness. Because the modified answers differ from the published ones, the expert wrote a new ground truth for each problem}. \hl{All five models were evaluated on the modified problems with the same prompting settings (zero-shot and CoT) and the same LLM judge as the main benchmark.

We compare each model's outcome (correct or incorrect) on the original and the modified version of each problem. Each comparison is one model on one problem in one prompting setting: 9 problems $\times$ 5 models gives 45 pairs per setting, for 90 comparisons across both settings.

The outcome was unchanged in 83.3\% of comparisons (75/90) and changed in 16.7\% (15/90). Of the changes, 6 went from incorrect to correct and 9 from correct to incorrect} (Table~\ref{tab:transform}). \hl{The pattern was similar in both settings, with 82.2\% of outcomes unchanged in zero-shot and 84.4\% in CoT. Overall accuracy moved only slightly, from 60.0\% to 56.7\%, and the difference is not statistically significant (exact McNemar test, $p \approx 0.61$).
}
\begin{table}[h]
\centering
\caption{Paired comparison of outcomes on original vs.\ modified problems (zero-shot and chain-of-thought combined).}
\label{tab:transform}
\begin{tabular}{lcccc}
\toprule
\textbf{Model} & \textbf{Same outcome} & \textbf{Changed} & \textbf{Incorrect $\rightarrow$ correct} & \textbf{Correct $\rightarrow$ incorrect} \\
\midrule
\modelicon{GPT}                & 16/18 (88.9\%) & 2/18 (11.1\%) & 2 & 0 \\
\modelicon{Claude_AI}      & 13/18 (72.2\%) & 5/18 (27.8\%) & 4 & 1 \\
 \modelicon{MistralAI}        & 15/18 (83.3\%) & 3/18 (16.7\%) & 0 & 3 \\
\modelicon{GeminiAI}  & 17/18 (94.4\%) & 1/18 (5.6\%)  & 0 & 1 \\
  \modelicon{Meta}   & 14/18 (77.8\%) & 4/18 (22.2\%) & 0 & 4 \\
\midrule
\textbf{Overall}      & \textbf{75/90 (83.3\%)} & \textbf{15/90 (16.7\%)} & \textbf{6} & \textbf{9} \\
\bottomrule
\end{tabular}
\end{table}

\hl{The models differ clearly. GPT-5 solved every modified problem, and Claude Sonnet~4.6's changes were mostly improvements (4 of 5). In contrast, every change for Mistral Large, Gemini~2.5 Flash-Lite and LLaMA~3.3~70B was a drop. These drops were concentrated in graph products, closed walks and connectivity. On those two problems, models that had solved the original version failed the modified one.

Overall, model rankings and accuracy levels were largely unchanged by the transformations. This suggests that our main findings are not driven mainly by memorisation. However, the problem-specific drops for the weaker models suggest they may partly rely on familiar problem templates.
}

\subsection{Group 3 --- Proof Construction}
\begin{figure}[ht]
  \centering
  \begin{subfigure}[t]{0.7\textwidth}
    \centering
    \includegraphics[width=\linewidth]{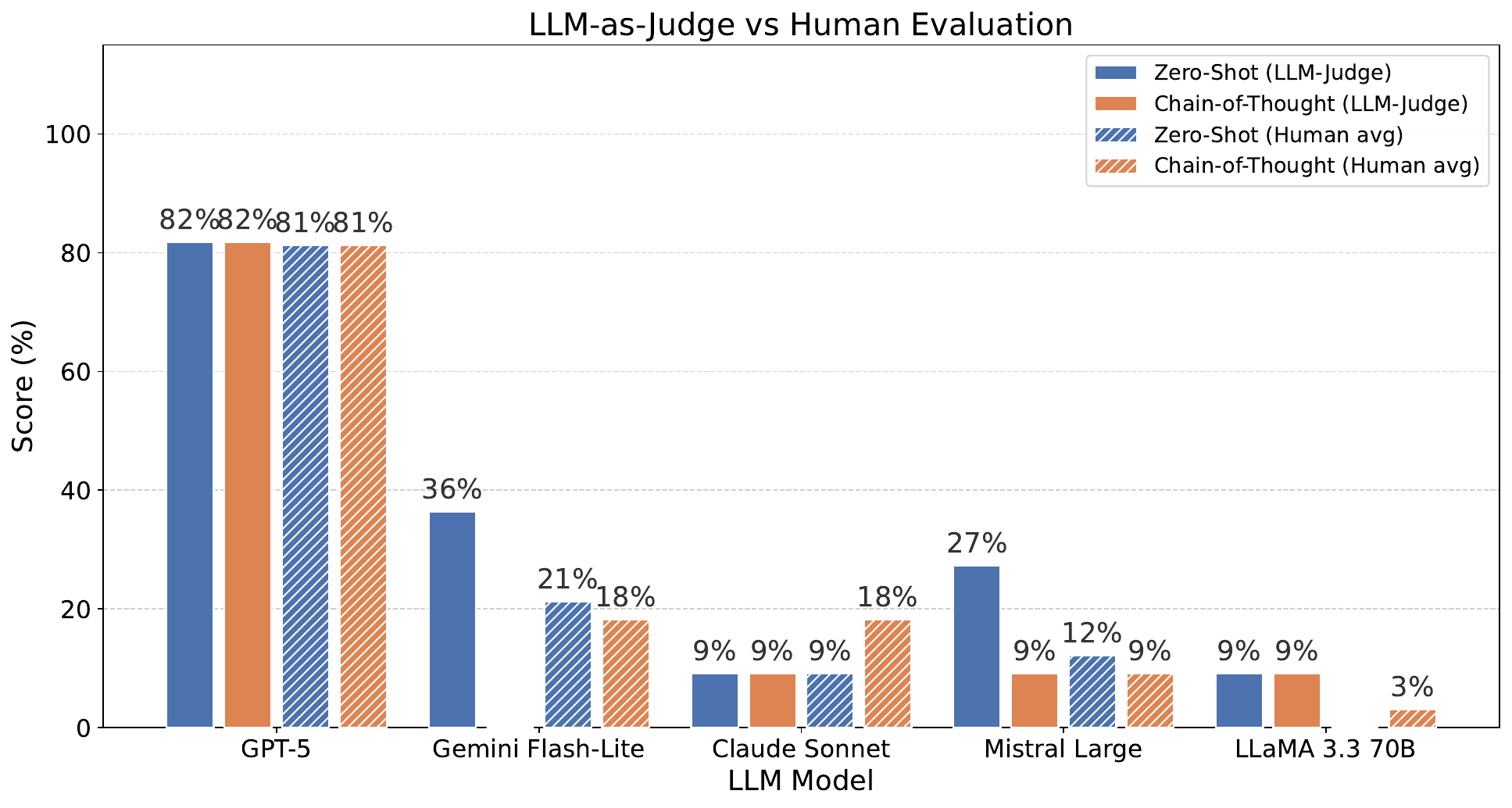}
  \end{subfigure}
  \caption{Comparison of Zero-Shot (ZS) and Chain-of-Thought (CoT) accuracy on Group 3
(proof-based problems) across five LLMs, evaluated by two methods: LLM-as-Judge
(solid bars) and human judge experts average (hatched bars).}
  \label{fig:main_results_3}
\end{figure}

Figure~\ref{fig:main_results_3} reports the accuracy of all
five models on Group~3 under zero-shot and CoT
prompting, evaluated by both the LLM judge (GPT-4o) and three
independent human experts (binarised scores \hl{where
each evaluator independently assigned a correct/incorrect
verdict based on their expert judgment of their own
three-point score.}). GPT-5 achieves
by far the strongest performance, with 82\% accuracy under
both conditions according to the LLM judge and 81\% under
human evaluation --- the only model for which automated and
human assessments are in close agreement. All other models
perform substantially below \texttt{GPT-5}: \texttt{Mistral Large~3} achieves
27\% (zero-shot, LLM judge) and 12\% (human), Claude
Sonnet~4.6 reaches 9\% under both evaluators on zero-shot,
and \texttt{Llama~3.3~70B} scores 9\% (LLM judge) and 0\% (human) ---
the weakest performance across the benchmark.

\texttt{Gemini~2.5 Flash-Lite} scores 36\% under the LLM judge
(zero-shot) and 21\% under human evaluation. The CoT
condition drops sharply to 0\% (LLM judge) and 18\% (human),
suggesting that CoT prompting amplifies
Gemini's verbosity to the point where neither the automated
judge nor human evaluators can reliably follow the argument.
Gemini produces substantially longer responses than all other
models, averaging 3{,}976 words per zero-shot answer and
7{,}611 words per CoT answer --- roughly 16$\times$ and
25$\times$ more verbose than \texttt{GPT-5} (243 and 310 words
respectively), and far exceeding the next most verbose model,
\texttt{Mistral Large~3} (856 and 1{,}424 words). Individual
zero-shot responses range from 2{,}002 to 5{,}900 words,
indicating that even Gemini's shortest answers are
substantially long. This extreme verbosity obscures valid
mathematical reasoning within lengthy outputs, confounding
both automated evaluation and human assessment: evaluators
reported difficulty identifying the core argument in
Gemini's responses, particularly under CoT where the
reasoning is further expanded. 

\noindent\textbf{\textit{Human vs.\ LLM judge.}}
A consistent pattern emerges across models: the LLM judge
and human experts largely converge for \texttt{GPT-5} (82\% vs.\
81\%) but diverge notably for weaker models. The gap is
most pronounced for \texttt{Gemini~2.5 Flash-Lite} under CoT (0\%
LLM vs.\ 18\% human), driven by answer length rather than
proof quality, as discussed above. For \texttt{Claude Sonnet~4.6},
the LLM judge and human evaluators agree on zero-shot
(9\% both) but diverge under CoT (9\% LLM vs.\ 18\%
human), suggesting that Claude's longer CoT responses
are partially credited by human experts but not by the
automated judge. For \texttt{Mistral Large~3}, the pattern reverses:
the LLM judge scores higher on zero-shot (27\% vs.\ 12\%
human), indicating that some answers the judge accepts as
correct are rejected by human experts as incomplete or
insufficiently rigorous. \texttt{Llama~3.3~70B} receives 0\% from
human experts on zero-shot --- the lowest human score in
the benchmark --- while the LLM judge assigns 9\%, again
suggesting the judge applies a more lenient standard on
short, plausible-sounding but ultimately incorrect proofs.
These divergences confirm that graduate-level proof
evaluation cannot be reliably delegated to an LLM judge
and that human assessment is essential for Group~3.

\noindent\textbf{\textit{Effect of CoT prompting.}}
CoT prompting produces mixed and largely negative results
on Group~3. \texttt{GPT-5} shows no change under CoT (82\% in both
conditions), suggesting it has reached its ceiling on these
problems. \texttt{Claude Sonnet~4.6} improves under human evaluation
from 9\% to 18\%, the only model to benefit consistently
from CoT at this level. \texttt{Mistral Large~3} degrades from 27\%
to 9\% (LLM judge) and from 12\% to 9\% (human), and
\texttt{Llama~3.3~70B} shows marginal improvement from 0\% to 3\%
(human) but no change under the LLM judge (9\% in both
conditions). \texttt{Gemini~2.5 Flash-Lite} collapses to 0\% under
the LLM judge and declines from 21\% to 18\% under human
evaluation, consistent with the verbosity amplification
effect described above. Overall, CoT prompting does not
provide the gains observed in Groups~1 and~2 for most
models, suggesting that structured step-by-step instructions
do not compensate for the fundamental limitations in
graduate-level mathematical reasoning at this stage of
model development.

\noindent\textbf{\textit{Inter-rater agreement.}}
Table~\ref{tab:kappa} reports Cohen's~$\kappa$ for all six
evaluator pairs across both prompting conditions.
Human--human agreement is heterogeneous: the E1--E2 pair
achieves substantial agreement under zero-shot ($\kappa =
0.77$) but drops to moderate under CoT ($\kappa = 0.48$), while E1--E3 achieves substantial agreement on
zero-shot ($\kappa = 0.69$) and excellent agreement
on CoT ($\kappa = 0.83$). The E2--E3
pair achieves moderate agreement across both conditions
($\kappa = 0.57$ ZS, $0.52$ CoT). Human--human agreement
averages $\kappa = 0.68$ (zero-shot) and $\kappa = 0.61$
(CoT), both meeting the minimum threshold of $\kappa > 0.60$
for acceptable agreement.
Human--LLM agreement is moderate across all three expert
pairs on zero-shot, with $\kappa$ ranging from $0.48$ to
$0.51$ (average $\kappa = 0.50$), indicating that the LLM
judge captures a meaningful signal but does not fully align
with any individual human expert. Agreement improves under
CoT for two of the three pairs: E2--LLM reaches $\kappa =
0.64$ (substantial) and E3--LLM improves to $\kappa = 0.55$
(moderate), while E1--LLM remains stable at $\kappa = 0.52$.

These results point to two distinct sources of disagreement.
First, the inherent subjectivity of graduate proof
evaluation --- even among domain experts, partial credit
boundaries are interpreted differently, as evidenced by
the E1--E2 drop from substantial to moderate agreement
under CoT ($\kappa = 0.77 \to 0.48$) and the residual
moderate E2--E3 agreement ($\kappa = 0.52$--$0.57$).
Second, the structural mismatch between the LLM judge's
strict binary verdict and the nuanced judgments of human
evaluators produces a systematic gap, most visible for
\texttt{Gemini~2.5 Flash-Lite} where response verbosity further
compounds the difficulty of automated assessment. Taken
together, these findings underscore a central limitation
of automated evaluation at the graduate level.
\begin{table}[h]
\centering
\caption{Inter-rater agreement (Cohen's $\kappa$) for
         Group~3 evaluation~\cite{landis1977measurement}.
         Landis \& Koch scale: $\kappa < 0.40$ = Fair;
         $0.40 \leq \kappa < 0.60$ = Moderate;
         $0.60 \leq \kappa < 0.80$ = Substantial;
         $\kappa \geq 0.80$ = Excellent.}
\label{tab:kappa}
\begin{tabular}{llcccc}
\toprule
\textbf{} & \textbf{Pair}
  & $\boldsymbol{\kappa}$ \textbf{(ZS)}
  & $\boldsymbol{\kappa}$ \textbf{(CoT)}
  & \textbf{Agreement (ZS)}
  & \textbf{Agreement (CoT)} \\
\midrule
\multirow{3}{*}{\rotatebox{30}{\textit{Human--Human}}}
 & E1--E2 & 0.77 & 0.48 & Substantial & Moderate        \\
 & E1--E3 & 0.69 & 0.83 & Substantial & Excellent   \\
 & E2--E3 & 0.57 & 0.52 & Moderate    & Moderate    \\
\midrule
\multirow{3}{*}{\rotatebox{30}{\textit{Human--LLM}}}
 & E1--LLM & 0.51 & 0.52 & Moderate & Moderate \\
 & E2--LLM & 0.48 & 0.64 & Moderate & Substantial \\
 & E3--LLM & 0.50 & 0.55 & Moderate & Moderate \\
\bottomrule
\end{tabular}
\end{table}

\noindent\textbf{\textit{Failure Topics.}}
Table~\ref{tab:tier3-failures} shows the failure mode
distribution across the 11 Group~3 topics. Failures are
not uniformly distributed --- certain topics consistently
draw incorrect responses from multiple models and multiple
evaluators, while others are handled correctly by most models.
The most broadly failed topic is \textit{odd cycles in
3-connected graphs}, where all four evaluators (E1, E2, E3,
and the LLM judge) flag failures across all five models.
Type~D (incomplete reasoning) dominates for human experts
and the LLM judge alike, indicating that models understand
the Fan Lemma argument in principle but consistently fail
to complete the construction of the four required odd cycles.
Similarly, \textit{maximum length paths sharing a vertex}
and \textit{matching saturating $X$ via degree condition}
are flagged by all three human experts and the LLM judge,
with Type~B dominating --- models apply the correct
strategy (pigeonhole or Hall's theorem) but introduce
logical errors during execution.
Type~A failures (hallucinated definitions) are rare in
Group~3 but appear exclusively on \textit{even cycles with
large average degree}, where E1 and E2 flag Type~A across
Claude, Gemini, Llama, and Mistral, indicating that these
models fabricate or misstate the bipartite subgraph theorem
used in the proof. This is the only topic where human
experts and the LLM judge disagree on the dominant failure
type: humans classify the errors as Type~A hallucinations
while the LLM judge classifies the same answers as Type~C
(wrong algorithm), consistent with the hypothesis that
the judge struggles to detect subtle definitional errors
in long-form proofs.
A systematic divergence between human and automated
evaluation appears on \textit{connectivity and
edge-connectivity} problems, where human experts flag
both Type~B and Type~D as co-dominant while the LLM
judge assigns only Type~B. Similarly, on \textit{long
paths from minimum degree}, the LLM judge identifies
Type~D (incomplete reasoning) as dominant while human
experts collectively flag Type~B. These divergences suggest
that models begin the correct argument --- often reaching
the key intermediate step --- but fail to complete the
final case analysis or induction, a distinction that human
evaluators are better positioned to identify.
Across all topics, failures are predominantly of Type~B
(correct algorithm, wrong execution), and in the majority
of cases at least two human evaluators independently flag
the same failure type, indicating consistent and
interpretable error patterns rather than isolated
disagreements.

\begin{table*}[ht]
\centering
\caption{Failure mode distribution by topic for Group~3
         (zero-shot condition, all five models combined).
         Each topic has two rows: the human row shows the
         judge label(s) of experts who flagged failures
         (E1, E2, E3); the LLM row shows the GPT-4o judge
         verdict.
         The dominant failure type per row is shown with a
         colored background.
         The rightmost column lists models with at least one
         incorrect answer on that topic:
         \raisebox{-2pt}{\includegraphics[height=9pt]{logos/gpt.png}}~\texttt{GPT-5},
         \raisebox{-2pt}{\includegraphics[height=9pt]{logos/Claude_AI.png}}~Claude,
         {\includegraphics[height=9pt]{logos/GeminiAI.png}}~Gemini,
         {\includegraphics[height=7pt]{logos/Meta.png}}~Llama,
         {\includegraphics[height=7pt]{logos/MistralAI.png}}~Mistral.
         \textcolor{textA}{\textbf{A}}~=~hallucinated definition;
         \textcolor{textB}{\textbf{B}}~=~correct algorithm, wrong execution;
         \textcolor{textC}{\textbf{C}}~=~wrong algorithm;
         \textcolor{textD}{\textbf{D}}~=~incomplete reasoning. $^\dagger$}
\label{tab:tier3-failures}
\resizebox{\textwidth}{!}{%
\begin{tabular}{
  p{3.8cm}
  l
  >{\centering\arraybackslash}p{0.9cm}
  >{\centering\arraybackslash}p{0.9cm}
  >{\centering\arraybackslash}p{0.9cm}
  >{\centering\arraybackslash}p{0.9cm}
  l
}
\toprule
\textbf{Topic}
  & \textbf{Judge}
  & \textcolor{textA}{\textbf{A}}
  & \textcolor{textB}{\textbf{B}}
  & \textcolor{textC}{\textbf{C}}
  & \textcolor{textD}{\textbf{D}}
  & \textbf{Models Failing} \\
\midrule

\multirow{2}{*}{\parbox{3.8cm}{\raggedright Maximum length paths:\\share a vertex}}
  & E1, E2, E3 & --- & \cellcolor{typeB}\textcolor{textB}{\textbf{X}} & --- & \textcolor{textD}{X} & \modelicon{Claude_AI} \modelicon{GeminiAI} \modelicon{Meta} \modelicon{MistralAI} \\
  & LLM   & --- & \cellcolor{typeB}\textcolor{textB}{\textbf{X}} & --- & \textcolor{textD}{X} & \modelicon{Claude_AI} \modelicon{Meta} \modelicon{MistralAI} \\
\midrule

\multirow{2}{*}{\parbox{3.8cm}{\raggedright Long paths from\\minimum degree}}
  & E1, E2, E3 & --- & \cellcolor{typeB}\textcolor{textB}{\textbf{X}} & \textcolor{textC}{X} & \textcolor{textD}{X} & \modelicon{Claude_AI} \modelicon{GeminiAI} \modelicon{Meta} \modelicon{MistralAI} \\
  & LLM   & --- & \textcolor{textB}{X} & \textcolor{textC}{X} & \cellcolor{typeD}\textcolor{textD}{\textbf{X}} & \modelicon{Claude_AI} \modelicon{GeminiAI} \modelicon{Meta} \modelicon{MistralAI} \\
\midrule

\multirow{2}{*}{\parbox{3.8cm}{\raggedright Even cycles: large\\average degree}}
  & E1, E2 & \cellcolor{typeA}\textcolor{textA}{\textbf{X}} & --- & \textcolor{textC}{X} & \textcolor{textD}{X} & \modelicon{Claude_AI} \modelicon{GeminiAI} \modelicon{Meta} \modelicon{MistralAI} \\
  & LLM   & --- & --- & \cellcolor{typeC}\textcolor{textC}{\textbf{X}} & --- & \modelicon{Claude_AI} \modelicon{GeminiAI} \modelicon{Meta} \modelicon{MistralAI} \\
\midrule

\multirow{2}{*}{\parbox{3.8cm}{\raggedright Eulerian circuit:\\consecutive edges}}
  & E1 & --- & \cellcolor{typeB}\textcolor{textB}{\textbf{X}} & \textcolor{textC}{X} & --- & \modelicon{gpt} \modelicon{Claude_AI} \modelicon{GeminiAI} \modelicon{Meta} \modelicon{MistralAI} \\
  & LLM   & --- & \textcolor{textB}{X} & \cellcolor{typeC}\textcolor{textC}{\textbf{X}} & --- & \modelicon{gpt} \modelicon{Claude_AI} \modelicon{Meta} \modelicon{MistralAI} \\
\midrule

\multirow{2}{*}{\parbox{3.8cm}{\raggedright Unique bipartition\\in bipartite graphs}}
  & E1, E2 & --- & \cellcolor{typeB}\textcolor{textB}{\textbf{X}} & --- & \textcolor{textD}{X} & \modelicon{Claude_AI} \modelicon{GeminiAI} \modelicon{Meta} \modelicon{MistralAI} \\
  & LLM   & --- & \cellcolor{typeB}\textcolor{textB}{\textbf{X}} & --- & --- & \modelicon{GeminiAI} \modelicon{Meta} \\
\midrule

\multirow{2}{*}{\parbox{3.8cm}{\raggedright Equal-length cycles\\(pigeonhole)}}
  & E1, E2, E3 & --- & \cellcolor{typeB}\textcolor{textB}{\textbf{X}} & --- & \textcolor{textD}{X} & \modelicon{Claude_AI} \modelicon{Meta} \modelicon{MistralAI} \\
  & LLM   & --- & \cellcolor{typeB}\textcolor{textB}{\textbf{X}} & --- & --- & \modelicon{Claude_AI} \modelicon{GeminiAI} \modelicon{Meta} \modelicon{MistralAI} \\
\midrule

\multirow{2}{*}{\parbox{3.8cm}{\raggedright Matching saturating\\$X$ via degree cond.}}
  & E1, E2, E3 & --- & \cellcolor{typeB}\textcolor{textB}{\textbf{X}} & \textcolor{textC}{X} & --- & \modelicon{Claude_AI} \modelicon{GeminiAI} \modelicon{Meta} \modelicon{MistralAI} \\
  & LLM   & --- & \cellcolor{typeB}\textcolor{textB}{\textbf{X}} & --- & \cellcolor{typeD}\textcolor{textD}{\textbf{X}} & \modelicon{Claude_AI} \modelicon{MistralAI} \\
\midrule

\multirow{2}{*}{\parbox{3.8cm}{\raggedright Connectivity \&\\edge-connectivity}}
  & E1, E2, E3 & --- & \cellcolor{typeB}\textcolor{textB}{\textbf{X}} & \textcolor{textC}{X} & \cellcolor{typeD}\textcolor{textD}{\textbf{X}} & \modelicon{Claude_AI} \modelicon{GeminiAI} \modelicon{Meta} \modelicon{MistralAI} \\
  & LLM   & --- & \cellcolor{typeB}\textcolor{textB}{\textbf{X}} & \textcolor{textC}{X} & --- & \modelicon{Claude_AI} \modelicon{GeminiAI} \modelicon{Meta} \\
\midrule

\multirow{2}{*}{\parbox{3.8cm}{\raggedright 2-connected graphs:\\paths via vertex}}
  & E1, E2 & --- & \cellcolor{typeB}\textcolor{textB}{\textbf{X}} & \textcolor{textC}{X} & \textcolor{textD}{X} & \modelicon{gpt} \modelicon{Claude_AI} \modelicon{GeminiAI} \modelicon{Meta} \modelicon{MistralAI} \\
  & LLM   & --- & \cellcolor{typeB}\textcolor{textB}{\textbf{X}} & \textcolor{textC}{X} & --- & \modelicon{Claude_AI} \modelicon{GeminiAI} \modelicon{Meta} \\
\midrule

\multirow{2}{*}{\parbox{3.8cm}{\raggedright Internally disjoint\\paths (2-connected)}}
  & E1, E2 & --- & \textcolor{textB}{X} & \cellcolor{typeC}\textcolor{textC}{\textbf{X}} & --- & \modelicon{Claude_AI} \modelicon{GeminiAI} \modelicon{Meta} \modelicon{MistralAI} \\
  & LLM   & --- & --- & \cellcolor{typeC}\textcolor{textC}{\textbf{X}} & --- & \modelicon{Claude_AI} \modelicon{Meta} \modelicon{MistralAI} \\
\midrule

\multirow{2}{*}{\parbox{3.8cm}{\raggedright Odd cycles:\\3-connected graphs}}
  & E1, E2, E3 & \textcolor{textA}{X} & \textcolor{textB}{X} & --- & \cellcolor{typeD}\textcolor{textD}{\textbf{X}} & \modelicon{gpt} \modelicon{Claude_AI} \modelicon{GeminiAI} \modelicon{Meta} \modelicon{MistralAI} \\
  & LLM   & --- & \textcolor{textB}{X} & \textcolor{textC}{X} & \cellcolor{typeD}\textcolor{textD}{\textbf{X}} & \modelicon{gpt} \modelicon{Claude_AI} \modelicon{GeminiAI} \modelicon{Meta} \modelicon{MistralAI} \\

\bottomrule
\end{tabular}%
}
\\
\begin{minipage}{\linewidth}
\vspace{4pt}
{\footnotesize $^\dagger$ \hl{Type~E (refusal) is omitted
as no model produced a refusal response in Group 3.}}
\end{minipage}
\end{table*}

\noindent\textbf{\textit{Implications for the learning
assistant use case.}}
From the perspective of a researcher using an LLM as a
self-study companion, Group~3 results carry a clear warning.
\texttt{GPT-5} is the only model that can reliably assist with
graduate-level graph theory proofs. \hl{GPT-5 achieves 82\% accuracy under the LLM
judge and 81\% under human expert evaluation on Group 3 zero-shot, the only model for which
automated and human assessments converge.} All other models produce
correct proofs fewer than one time in three, and in several
cases --- \texttt{Llama~3.3~70B} (0\% human, zero-shot) and Mistral
Large~3 (12\% human, zero-shot) --- correct answers are
vanishingly rare. A student relying on these models for
proof guidance would receive incorrect or incomplete arguments the majority of the time, with no reliable mechanism to detect the error. \hl{This pattern is not specific to graph theory, but reflects a structural property of current LLMs on formal mathematical reasoning tasks that has direct implications for the design of AI-assisted research workflows.}


\section{Conclusion}
\label{sec:conclusion}

This paper introduced \texttt{GTBench}, a curriculum-grounded
benchmark for evaluating the graph-theoretic reasoning
capabilities of LLMs as self-study tools for researchers and
students. Spanning 63 problems across three groups --- from
undergraduate definitions to graduate proof construction ---
\texttt{GTBench} was designed to mirror the natural progression
of graph theory instruction and to assess whether LLMs can
serve as a reliable learning assistant at each stage of that
progression. Five frontier models were evaluated under
zero-shot and chain-of-thought prompting, with automated
LLM-as-judge evaluation for Groups~1 and~2 and human expert
evaluation combined with LLM-as-judge for Group~3. Our
results reveal a clear and consistent hierarchy: \texttt{GPT-5}
dominates across all groups and conditions, approaching
ceiling performance on Group~1 and maintaining meaningful
accuracy on graduate-level proofs. All other models degrade
substantially as problem difficulty increases, with
\texttt{Llama~3.3~70B} achieving 0\% under human evaluation on
Group~3 zero-shot. \hl{Model errors are also partly reproducible: 54.9\% of incorrect zero-shot responses on Group~1 and 70.4\% on Group~2 were incorrect again in all three repeated runs, so re-querying a model is not a reliable way for a user to detect its errors, particularly on algorithmic problems.} \hl{A transformation test on expert-modified variants of nine problems left 83.3\% of outcomes unchanged, suggesting that these results are not driven mainly by memorisation of published solutions.}


Beyond accuracy, our failure mode analysis reveals that
Type~B errors (correct algorithm, wrong execution) dominate
Groups~1 and~2, while graduate-level problems in Group~3
additionally surface Type~D failures (incomplete reasoning)
and, in several topics, genuine disagreement between human
evaluators --- a finding that itself confirms the inherent
difficulty of automated proof assessment at this level.
The inter-rater agreement observed in Group~3 ($\kappa$
ranging from $0.48$ to $0.83$ across human pairs)
underscores that graduate proof evaluation resists
standardisation, and that the LLM judge, despite achieving
moderate agreement with individual experts ($\kappa =
0.48$--$0.64$ on Group~3), systematically underestimates model
performance on verbose or near-complete proofs. 
\hl{Taken together, these findings carry a practical conclusion
for the AI-enabled scientific research community: current
LLMs can plausibly support self-directed learning in graph
theory at the introductory and intermediate levels, but
graduate-level proof construction requires human expert
verification --- none of the evaluated models should be trusted as a
sole guide at this level of mathematical reasoning.
\texttt{GTBench} provides a replicable foundation for
tracking these limitations as model capabilities evolve,
and for informing the design of AI-assisted research
workflows that account for the structured,
difficulty-dependent nature of LLM failure in formal
mathematical domains.}

\textbf{\textit{Future Work.}}
While \texttt{GTBench} establishes a rigorous foundation for evaluating graph-theoretic reasoning in LLMs, several limitations point to natural extensions. The current 63-problem scope, though carefully graded, restricts generalizability; expanding coverage to adjacent mathematical domains — combinatorics, linear algebra, abstract algebra, and topology — would clarify whether the sharp performance degradation observed at the graduate level is domain-specific or reflects a fundamental ceiling in formal mathematical reasoning across frontier models.
Evaluation methodology represents the most immediate direction for improvement. The current three-point scoring scale (0, 0.5, 1) is a coarse approximation of proof quality that collapses meaningful distinctions — a response that correctly states all lemmas but fails only the final deduction is scored identically to one that fails entirely after the first step. Future work should replace this scheme with a continuous percentage-based scoring rubric that awards partial credit proportionally across well-defined proof dimensions: problem comprehension, strategy selection, intermediate reasoning, and final execution. \hl{Critically, both human evaluators and the LLM judge
should operate on the same continuous scale from the
outset, eliminating the post-hoc binarisation step
required in the current study to align the two
evaluation protocols.} Such granularity would expose precisely where in the reasoning chain models break down, enable finer-grained comparison across models and difficulty tiers, and provide a more honest signal for practitioners deciding when to trust LLM-generated mathematical arguments. 
\hl{The partial reproducibility of incorrect responses (54.9--70.4\% on Group~1 and Group~2) raises the question of why some errors persist while others disappear on re-sampling. Understanding whether persistent errors originate from pretraining data gaps, distributional biases, or architectural properties is a prerequisite for designing systems in which errors are detectable rather than confidently invisible.}
\section*{Acknowledgment}
P. Diehl was supported by the U.S. Department of Energy through the Los Alamos National Laboratory. Los Alamos National Laboratory is operated by Triad National Security, LLC, for the National Nuclear Security Administration of U.S. Department of Energy (Contract No. 89233218CNA000001). Approved by Los Alamos National Laboratory as LA-UR-26-24106.

\section*{Data Availability}
The LLM-generated solutions for all problems and all
models, along with the benchmark JSON files and scoring
sheets, will be made publicly available on
 Zenodo~\cite{} upon acceptance of this paper. 

\section*{Code Availability}
The source code will be made publicly available on
GitHub\footnote{\url{}} and
Zenodo~\cite{} upon acceptance of this paper. 

\bibliographystyle{plain}
\bibliography{software}
\end{document}